\documentclass[10pt,twocolumn,letterpaper]{article}

\usepackage[pagenumbers]{cvpr} 

\usepackage[sort&compress,numbers]{natbib}
\usepackage{mathrsfs}
\usepackage{amsmath}
\usepackage{bm}
\usepackage{multirow}
\usepackage{adjustbox}
\usepackage{graphicx}
\usepackage{colortbl}
\usepackage{microtype} 
\usepackage{amsmath, amssymb, mathtools}
\usepackage{appendix}
\usepackage{wrapfig}
\usepackage[pagebackref,breaklinks,colorlinks,citecolor=cvprblue]{hyperref}
\usepackage{bbm}

\def\paperID{929} 
\def\confName{CVPR}
\def\confYear{2024}

\definecolor{mygray}{gray}{0.6}
\definecolor{cvprblue}{rgb}{0.21,0.49,0.74}

\newcommand{\OurMethod}{DIM}

\title{Discover and Mitigate Multiple Biased Subgroups in Image Classifiers}

\author{
\thanks{Equal contribution.}~~Zeliang Zhang \quad $^*$Mingqian Feng \quad \thanks{Project lead. Work done before Zhiheng Li joining Amazon.}~~Zhiheng Li \quad Chenliang Xu\\
University of Rochester\\
{\tt\small \{zeliang.zhang,mingqian.feng,zhiheng.li,chenliang.xu\}@rochester.edu}
}

\begin{document}
\maketitle

\begin{abstract}
    Machine learning models can perform well on in-distribution data but often fail on biased subgroups that are underrepresented in the training data, hindering the robustness of models for reliable applications. Such subgroups are typically unknown due to the absence of subgroup labels. Discovering biased subgroups is the key to understanding models' failure modes and further improving models' robustness. Most previous works of subgroup discovery make an implicit assumption that models only underperform on a single biased subgroup, which does not hold on in-the-wild data where multiple biased subgroups exist.
    
    In this work, we propose Decomposition, Interpretation, and Mitigation (\OurMethod{}), a novel method to address a more challenging but also more practical problem of discovering multiple biased subgroups in image classifiers. Our approach decomposes the image features into multiple components that represent multiple subgroups. This decomposition is achieved via a bilinear dimension reduction method, Partial Least Square (PLS), guided by useful supervision from the image classifier. We further interpret the semantic meaning of each subgroup component by generating natural language descriptions using vision-language foundation models. Finally, \OurMethod{} mitigates multiple biased subgroups simultaneously via two strategies, including the data- and model-centric strategies. Extensive experiments on CIFAR-100 and Breeds datasets demonstrate the effectiveness of \OurMethod{} in discovering and mitigating multiple biased subgroups. Furthermore, \OurMethod{} uncovers the failure modes of the classifier on Hard ImageNet, showcasing its broader applicability to understanding model bias in image classifiers. The code is available at \url{https://github.com/ZhangAIPI/DIM}.
\end{abstract}
\vspace{-2em}

\section{Introduction}


Machine learning models can achieve overall good performance on in-distribution data~\citep{huang2023egocentric,jiang2023one}. However, they often underperform on certain biased subgroups that are underrepresented in the training data, undermining model robustness against group distributional shifts~\cite{sagawa2019distributionally}. For instance, ResNet~\citep{he2016deep} trained on ImageNet~\cite{deng2009imagenet} fails to recognize the  ``balance beam'' when kids are not present,  where the image classifier is biased towards the subgroup of ``balance beam'' when kids are present (\cf \cref{fig: hard_imagenet_exp}). Therefore, identifying and mitigating subgroup biases in image classifiers is crucial to improve models' reliability and robustness~\citep{kaur2022trustworthy, du2022towards,li2023trustworthy,deng2009imagenet,luo2023zero,zhang2024bag,zhang2024random}.

\begin{figure}[t]
    \centering
    \includegraphics[width=\linewidth]{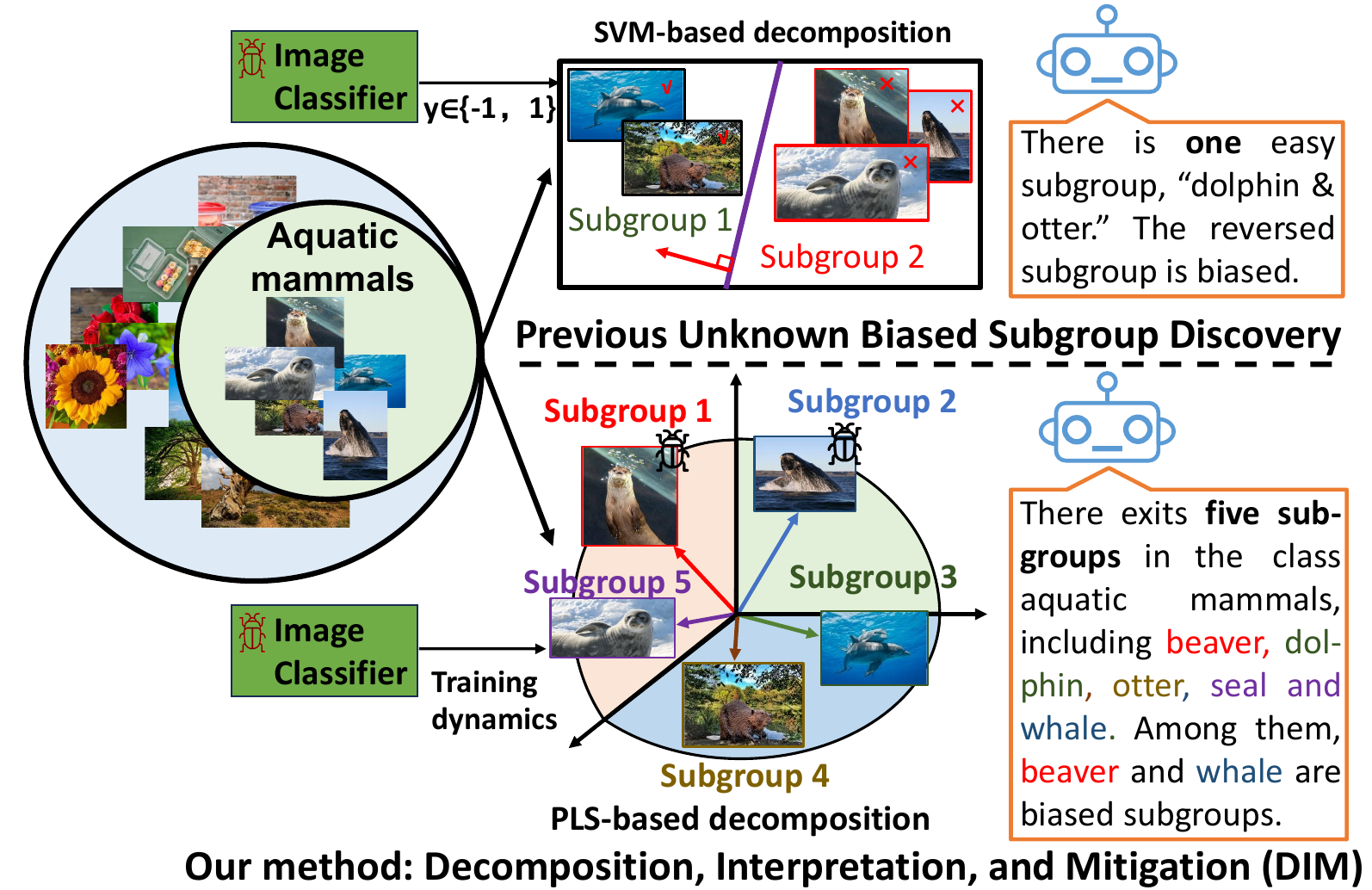}
    \caption{While the previous method~\citep{jain2022distilling} exploits the SVM to detect the single biased subgroup using the classification correctness on samples, we propose to integrate the training dynamics of biased image classifiers as the supervision into PLS decomposition to discover multiple unknown subgroups. This allows us to further subtly interpret discovered subgroups and precisely mitigate biases.}
    \label{fig: motivation}
    \vspace{-4mm}
\end{figure}

However, previous research on model bias has many limitations. \underline{First}, most existing methods~\citep{karkkainen2021fairface, zhao2022towards,caliskan2022gender,wang2022revise} for bias discovery {require structured attributes}, demanding expensive human labor to collect and label data. The impracticality of annotating every possible attribute also leads to potential biases remaining hidden~\citep{li2021discover}. \underline{Second}, few works involve mining the unknown multi-bias. While \citet{eyuboglu2022domino} approach this problem by employing a Gaussian Mixed Model~\citep{banfield1993model} to detect biases through clustering, their method relies on proxies of clusters to explain unknown biases, which can result in interpretation distortion (see the Domino part of \cref{fig:  retrieval_cifar} in \cref{sec: exp_cifar}). To address this issue, recent work by \citet{jain2022distilling} proposes directly distilling failure modes as firsthand directions in latent feature space and using cross-modal representation~\citep{zhen2019deep} for bias interpretation. However, this method only detects a single shortcut and falls short in multi-shortcut discovery. Compared with single bias, the discovery of multi-bias is more practical and challenging~\citep{witzgall2022reducing, li2023whac}. The lack of efficient bias discovery techniques impedes the application of bias mitigation methods, which typically require the knowledge of bias attributes~\citep{sagawa2019distributionally,wang2020towards}. Thus, efforts on multiple unknown biases are essential and challenging to enable practical and scalable solutions to trustworthy AI. 


In this paper, we are motivated by the insight that a single unknown bias can be identified in the latent space~\citep{jain2022distilling}, and then move forward to addressing a more challenging problem, \textit{multiple unknown biased subgroups}. Although models are designed to grasp predefined classes (labels) during training, they also inadvertently learn subgroups distinguished by unconscious attributes. The distribution difference between subgroups and the presence of multi-bias result in image classifiers exhibiting erratic performance across subgroups. As shown in \cref{fig: motivation}, the class of ``aquatic mammals'' in the CIFAR-100 dataset~\citep{krizhevsky2009learning} can be further divided into $5$ subgroups, ``beaver,'' ``dolphin,'' ``otter,'' ``seal,'' and ``whale.'' While a ResNet-18~\citep{he2016deep}  has an overall accuracy of $54.6\%$ on this ``aquatic mammals'' class, it performs poorly on ``beaver'' ($34\%$) and ``whale'' ($37\%$). This disparity manifests the existence of subgroup bias and its impairment of the robustness of the model. Limited to simply using sample correctness, \citet{jain2022distilling} can discover only two broad subgroups: those with superior and inferior performance. Specifically, it discerns the positive group of two well-performance subgroups, ``dolphin'' and ``otter,'' through a unified direction in latent space. However, the opposite direction, supposed to embody biased subgroups (\textit{i.e.}, ``beaver'' and ``whale''), fails to maintain representativeness and interpretability (see the \citet{jain2022distilling} part of \cref{fig:  retrieval_cifar} in \cref{sec: exp_cifar}).

Addressing the problem of \textit{multiple unknown biased subgroups} presents three challenges. The first difficulty lies in the lack of explicit supervisory signals from the model to guide the discovery process. The second hurdle is interpreting identified directions and pinpointing biased ones among them. Third, beyond the identification and interpretation, there is a consequential line of inquiry into how these discerned subgroups can be harnessed for bias mitigation.

In response to these challenges, we propose an innovative \textit{latent space-based multiple unknown biased subgroup discovery} method, named \textbf{\OurMethod{}} (\textbf{D}ecomposition, \textbf{I}nterpretation, and \textbf{M}itigation). Initially, we integrate the training dynamics of the biased model into the partial least squares (PLS)~\citep{abdi2013partial} to supervise the decomposition of image features in latent space. Those features are decomposed into different subgroup directions, each aligning with different subgroups that the image classifier learned. Subsequently, these directions are utilized to generate pseudo-subgroup labels for the original dataset to distinguish biased subgroups that exhibit lower accuracy. Finally, upon identifying multiple subgroups, including biased ones, \OurMethod{} employs cross-modal embeddings in the latent space to interpret these subgroups and annotate the data with subgroup information to mitigate biases.

Our work presents three key contributions as follows,
\begin{enumerate}
    \item We formulate the problem of discovering multiple unknown biased subgroups and subsequent bias mitigation.
    \item We propose \textbf{\OurMethod} (\textbf{D}ecomposition, \textbf{I}nterpretation, and \textbf{M}itigation), a novel framework used to discover, understand, and mitigate multiple unknown biased subgroups learned by image classifies.
    \item We conduct experiments on three datasets: CIFAR-100, Breeds, and Hard ImageNet. We verify \OurMethod's ability to detect biased subgroups on CIFAR-100 and Breeds, where classes and ground-truth subgroups, including biased ones, are given. For Hard ImageNet, we apply \OurMethod{} to discover biased subgroups implicitly learned by the image classifier, thereby illustrating its failure modes.
\end{enumerate}



\section{Related Work}
\label{sec: related_work}

\noindent \textbf{Bias Identification} \quad  Many works have been proposed to identify and explain the bias of deep learning models. \citet{eyuboglu2022domino} leverage cross-modal embeddings and a novel error-aware model to discover underperforming slices of samples. \citet{singla2022salient} use the activation maps for neural features to highlight spurious or core visual features and introduce an ImageNet-based dataset, Salient ImageNet, which contains masks of core visual and spurious features. 
\citet{zhu2022gsclip} propose a training-free framework, GSCLIP, to explain the dataset-level distribution shifts. \citet{li2021discover} and \citet{lang2021explaining} use a generative model to discover and interpret unknown biases. \citet{jain2022distilling} harness the linear classifier to identify models' failure mode and uses CLIP \citep{radford2021learning} for the automatic caption to explain the failure mode. Previous methods usually fall on the single-shortcut problems. However, real-world scenes usually involve multiple biases, including multiple subgroup biases, posing challenges to existing methods. Our work proposes a novel approach to identify and explain multiple unknown subgroup biases, which is scalable to large real-world datasets.

\noindent \textbf{Bias Mitigation} \quad There are many approaches proposed to mitigate the bias, such as the re-sampling and weighting strategy~\cite{li2019repair,qraitem2023bias}, distributional robust optimization~\cite{slowik2022distributionally,wen2022distributionally}, invariant risk minimization~\cite{mao2023debiasing}, and adversarial debiasing~\cite{zhang2018mitigating,lim2023biasadv}. Some work also exploits the identified bias to mitigate the bias of models, such as EIIL~\cite{creager2021environment} and LfF~\cite{NEURIPS2020_eddc3427}. However, these methods require the bias labels in the training dataset, which are usually unknown in practice, leading to poor scalability. To address this issue, some work has proposed research on bias mitigation without access to bias annotation. \citet{nam2020learning} propose to train a debiased model on samples, which is against the prejudice of the well-trained bias model. \citet{pezeshki2021gradient} propose a regularization term to decouple failure learning dynamics. \citet{li2022discover} propose the debiasing alternate networks to discover unknown biases and unlearn the multiple identified biases for the classifier. \citet{park2023training} propose the debiased contrastive weight pruning to investigate unbiased networks. In our work, we employ identified bias subgroup information to implement data-centric and model-centric strategies for bias mitigation, improving the model's robustness.

\section{Problem Formulation}
\label{sec: task_def}

Consider an image classifier trained on the data $\bm{x} \in \mathcal{X}$ with annotated ground-truth label $y \in \mathcal{Y}$. In this context, assume there are $L$ annotated classes in the dataset, making $\mathcal{Y}=\{1,2,...,L\}$. We hypothesize each class contains $G$ subgroups\footnote{While the number of subgroups \(G\) might vary across classes, for simplicity, we consider an equal number of subgroups in each class.} that share certain characteristics or features, a total of $L \times G$ subgroups. For each input $\bm{x}$, we denote its subgroup membership as $g \in \{1,2,..., G\}$. The input data, labels, and subgroups are drawn from a joint distribution $P$.


An ideal classifier without subgroup bias should maintain consistent performance across all $G$ subgroups. Conversely, a biased classifier is characterized by its inferior performance on specific subgroups. The primary objective is twofold: first, to discover a total of $L \times G$ subgroups, and second,  to identify the biased $k$ subgroups (out of $G$) in each class, which exhibit lower classification accuracy than the median. The insight that mitigating one bias in an image classifier with multiple biases can inadvertently amplify others \citep{li2023whac} underscores the importance of addressing multiple unknown biases. Hence, we particularly focus on multiple unknown biased subgroups ($k \geq 2$), which implies $G \geq 4$.

Following the identification, the subsequent tasks involve interpreting and mitigating these biases. This progression is key to not only understanding but also enhancing the robustness of image classifiers against subgroup biases.

\begin{figure*}[t]
    \centering
    \includegraphics[width=0.82\linewidth]{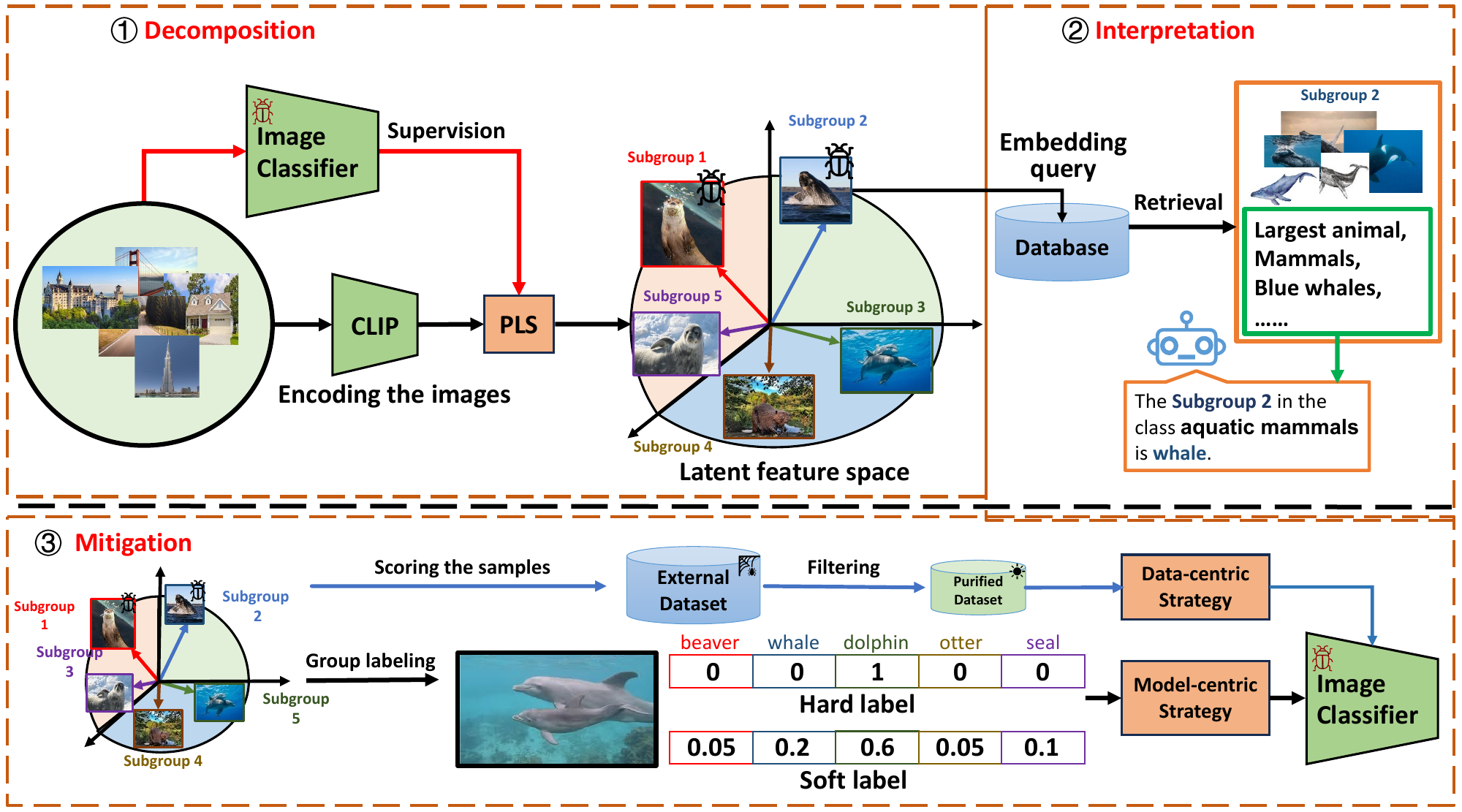}
    \caption{Overview of \OurMethod{} method. \OurMethod{} consists of three stages: Decomposition, Interpretation, and  Mitigation. At the decomposition stage, we decompose the image features of ``aquatic mammals'' into the embedding directions of multiple subgroups. Then, we interpret the discovered subgroup embeddings with text descriptions, \eg, the subgroup ``whale'' in class ``aquatic mammals.'' At the mitigation stage, we propose data-centric and model-centric strategies to mitigate the subgroup bias to improve the robustness of the image classifier.}
    \label{fig: pipeline}
    \vspace{-3mm}
\end{figure*}

\section{Method}
\label{sec: method}


In the quest to address multiple unknown biased subgroups, we propose a novel method, \textbf{DIM} (\textbf{D}ecomposition, \textbf{I}nterpretation, and \textbf{M}itigation), as shown in \cref{fig: pipeline}.


\label{sec: dim}

\subsection{Decomposition}

\noindent \textbf{Discovering Multiple Subgroups.} The previous study~\citep{lin2022zin} has identified that \textit{it is theoretically impossible to derive invariant features from the heterogeneous data without environment information}~\citep{rohrer2018thinking,liu2023towards}. This implies that without any additional information, unsupervised biased subgroup discovery is also impossible. Motivated by this, we integrate the model supervision into the partial least square (PLS)~\citep{abdi2013partial} method to decompose the image features into multiple subgroup directions in the latent space. Unlike the principal component analysis (PCA)~\citep{jolliffe2016principal}, which is unsupervised decomposition, PLS can be used to derive the components within the inputs mostly aligned with the supervision.

Concretely, there are three steps to discovering multiple unknown subgroups in a specific class (fixing $y\!=\!l$). Initially, we use the CLIP image encoder, a function $f_{\text{image}} \colon \mathcal{X} \rightarrow \mathbb{R}^d$, where $d$ denotes the latent feature dimension, to encode images in this class, which can be considered as a variable that follows specific empirical class distribution $\hat{P}_l$. The images are encoded into input embeddings $\widehat{\bm{x}} \coloneqq f_{\text{image}}(\bm{x})$ in the latent space. Next, for each image $\bm{x}$, we collect some information provided by the studied model, denoted as $\bm{z_x} \in \mathbb{R}^{M}$, where $M$ is the number of used information. The information, including the loss, correctness, logit, \textit{etc.}, serves as supervision guiding the decomposition. Subsequently, we apply the PLS method to model the decomposition of $\widehat{\bm{x}}$ supervised by $\bm{z_x}$. The core idea is to search for subgroup directions in the latent space that maximize the correlation with principal components in the supervision. In discovering the first subgroup, $i=1$, setting $\widehat{\bm{x}}_1 \coloneqq \widehat{\bm{x}}$ and $\bm{z}_{\bm{x}, 1} \coloneqq \bm{z}_{\bm{x}}$, it can be formulated as the following optimization target,
\begin{equation}
    \setlength\abovedisplayskip{3pt} 
    \setlength\belowdisplayskip{3pt}
    \begin{aligned} \label{eq: PLS}
    &\max_{\bm{w}_{i},~\bm{h}_{i}} \mathbb{E}_{\hat{P}_l}(u_i v_i)=\max_{\bm{w}_{i},~\bm{h}_{i}} \mathbb{E}_{\hat{P}_l}\left((\bm{w}_{i}^T \widehat{\bm{x}}_i)(\bm{h}_{i}^T \bm{z}_{\bm{x},i})\right),
    \end{aligned}
\end{equation}
where $\bm{w}_{i} \in \mathbb{R}^d$, $\bm{h}_{i} \in \mathbb{R}^M$ are the normalized directions and $u_i \coloneqq \bm{w}_{i}^T \widehat{\bm{x}}_i$, $v_i \coloneqq \bm{h}_{i}^T \bm{z}_{\bm{x}, i}$ represent the latent scores on the directions. For each image, a high latent score implies a high similarity to the discovered subgroup. Upon the optimization, $\widehat{\bm{x}}_i$ and $\bm{z}_{\bm{x}, i}$ are updated by subtracting the discovered information. It can be achieved by regressing $\widehat{\bm{x}}_i$ on $u_i$,and regressing $\bm{z}_{\bm{x},i}$ on $v_i$,
\begin{equation}
    \setlength\abovedisplayskip{3pt} 
    \setlength\belowdisplayskip{3pt}
         \widehat{\bm{x}}_i=u_i\bm{\alpha}_i+\bm{\epsilon}_i,~\bm{z}_{\bm{x},i}=v_i\bm{\beta}_i+\bm{\eta}_i,
\end{equation}
where $\bm{\alpha}_i$, $\bm{\beta}_i$ are regression coefficients and $\bm{\epsilon}_i$, $\bm{\eta}_i$ are the remainders and then setting $\widehat{\bm{x}}_{i+1}=\bm{\epsilon}_i$ and $\bm{z}_{\bm{x},i+1}=\bm{\eta}_i$.

By iteratively repeating the optimization and the update $n$ time, we decompose class input embeddings $\widehat{\bm{x}}$  as $\widehat{\bm{x}}=\sum_{i=1}^n u_i\bm{\alpha}_i+\bm{\epsilon_{n+1}}$ and obtain a set of discovered subgroup vectors $\{\bm{w}_i\}_1^n$, where $n$ is a hyperparameter.

\noindent \textbf{Identifying Biased Subgroups.} Following discovering subgroup directions, the next step is to pinpoint further which of them are biased. We achieve this by computing pseudo-subgroup labels for images in a held-out validation set and evaluating the model's performance on each discovered subgroup. For each validation image, we compute the subgroup score $\bm{u}\coloneqq(u_1,...,u_n)^T\in \mathbb{R}^n$, assigning the subgroup with the highest score as the pseudo-subgroup label. In the soft-label case, we directly use the subgroup score $\bm{u}$. Subsequently, we compute the model's accuracy on images with each pseudo-subgroup label and identify the $k$ subgroups with the top-$k$ worst accuracy as $k$ biased subgroups.

\subsection{Interpretation}

Interpreting the discovered subgroups is a critical step to bridge the gap between abstract representations and meaningful insights~\citep{dunlap2023describing}. The ultimate goal is to generate natural language descriptions of biased subgroups. However, the discovered subgroup embeddings are not directly interpretable. Thus, we leverage the retrieval approach to interpret the discovered subgroup, as shown in the second part of \cref{fig: pipeline}. For each subgroup, the retrieval results are pairs of images and texts (metadata, including descriptions) from the LAION-5B dataset~\citep{schuhmann2022laion}, providing visual illustration and contextual information. To further refine the understanding, we collect all retrieved descriptions and utilize a large language model (LLM) for summarization to explain the subgroup.

\subsection{Mitigation}
In \OurMethod{}, we propose data-centric and model-centric strategies using the discovered subgroups for bias mitigation.




The \textbf{data-centric strategy} is applied to scenarios where access to a substantial external dataset is available. The goal is to enhance the model's performance on biased subgroups by adding a limited number of samples to the training set (due to computational constraints). To achieve this, we leverage the discovered subgroup embeddings to filter extra data to identify high-quality samples that show a strong correlation with the bias subgroups. For each image in the external pool, we first compute its subgroup scores, specifically on those bias subgroups. We then select images with the highest subgroup scores for each biased subgroup. By deliberately increasing the representation of these subgroups in the training set, we aim to systematically mitigate the multiple unknown subgroup biases in the image classifier.


The \textbf{model-centric strategy} is proposed to leverage the discovered subgroups to annotate images in the training set with pseudo-subgroup labels and integrate those labels into existing supervised bias mitigation methods. While the hard label, obtained by taking the argmax of subgroup score $\bm{u}=(u_1,...,u_n)^T$, can be directly applied to supervised mitigation methods, we also capitalize on the soft label to improve generalization. The motivation is that one image may contain multiple biases, making it belong to multiple subgroups. Thus, some supervised mitigation methods, such as the groupDRO~\citep{sagawa2019distributionally} and DI~\citep{wang2020towards}, are relaxed to the soft-label version. We provide an example of adapting DI to soft-label (Soft-DI) as follows. More details for the soft-label version of mitigation methods can be found in \cref{app:alg_soft}. 

\noindent \textbf{\textit{Case study of model-centric strategy with Soft-DI.}} \quad We consider Domain Independent (DI)~\citep{wang2020towards} method with $G$ domains (\ie, number of subgroups), which contains $G$ classification heads sharing features extracted by the backbone. In the original DI, when training on the data $\bm{x}$ with known hard group label $g \in \{1,2,...,G\}$, the model's output is $\hat{\bm{p}} = \hat{\bm{p}}_g$, where $\hat{p}_g$ is the $g$-th classifier's output.
For the soft-label case, where $\bm{g} \! = \! (g_1,...,g_G) \! \in \! \mathbb{R}^G$, we define the training output as $\hat{\bm{p}} = \sum_{i=1}^G g_i \hat{\bm{p}}_i$. When performing inference on the test set without group annotation, soft-DI maintains the original method so that $\hat{\bm{p}} = \frac{1}{G} \sum_{i=1}^G \bm{p}_i$.

\section{Experiments}
\label{sec: exp}


\subsection{Setup} \label{sec: setup}
\noindent \textbf{Datasets}. Our experiment uses three datasets, including CIFAR-100~\citep{krizhevsky2009learning}, Breeds~\citep{santurkar2021breeds}, and Hard ImageNet~\citep{moayeri2022hard}.
\begin{itemize}
\item The CIFAR-100~\citep{krizhevsky2009learning} dataset contains $100$ fine-classes, each with $500$ images for training and $100$ images for testing. These fine-classes are organized into 20 superclasses, with every $5$ fine-classes constituting one superclass. To clarify, from now on, we treat superclasses as classes and fine-classes as the ground-truth subgroups.
\item Breeds~\citep{santurkar2021breeds}, a subset of the ImageNet-1K, is composed of $130$ fine-classes. Every $10$ fine-classes is grouped into one class, resulting in a total of $13$ classes. 
\item Hard ImageNet~\citep{moayeri2022hard}, another subset of the ImageNet-1K, consist of $15$ classes, without further subdivision into finer classes. This dataset is particularly challenging due to strong spurious correlations. The absence of ground truth for discovering multiple unknown subgroups offers a valuable testbed for our method. We apply \OurMethod{} to mine multiple unknown biases and delineate models' failure. 
\end{itemize}

\noindent \textbf{Baselines}. In our experiments, we evaluate three tasks: the discovery of multiple unknown subgroups, the identification of biased subgroups, and the mitigation of pinpointed biases. For discovery and identification tasks, our method, Decomposition, Interpretation, and Mitigation (\OurMethod), is compared with \citet{jain2022distilling}, Domino~\citep{eyuboglu2021domino}. We also test our method by replacing PLS with PCA (\OurMethod-PCA) as an ablation study attesting to the importance of supervision during the decomposition phase. For the mitigation task, we annotate the samples with pseudo labels generated by \citet{jain2022distilling} and ours. These labels are then applied to various mitigation methods to verify the effectiveness of discovered subgroups on bias reduction. We implement 1) unsupervised model-centric methods: JTT~\citep{liu2021just}, SubY~\citep{idrissi2022simple}, LfF~\citep{nam2020learning}, EIIL~\citep{creager2021environment}; 2) supervised model-centric methods: groupDRO (gDRO)~\citep{sagawa2019distributionally}, soft-label groupDRO (Soft-gDRO), DI~\citep{wang2020towards}, and soft-label DI (Soft-DI), and 3) the data-centric strategy by intervention through filtering extra data~\citep{jain2022distilling}, with the decision value of \citet{jain2022distilling} and ours.

\noindent \textbf{The selection of supervision}. Supervision plays a crucial role in guiding the decomposition within the latent feature space in our methodological design. We adopt $3$ distinct training dynamics as supervision: 1) correctness: whether the model accurately classifies a given image; 2) logit: the logit output of the model on each image; and 3) loss: the value of loss function for each image. Besides, we also utilize the ground-truth subgroup features as an additional form of supervision for a comprehensive evaluation. 

\begin{figure*}
\centering
    \includegraphics[width=0.95\textwidth]{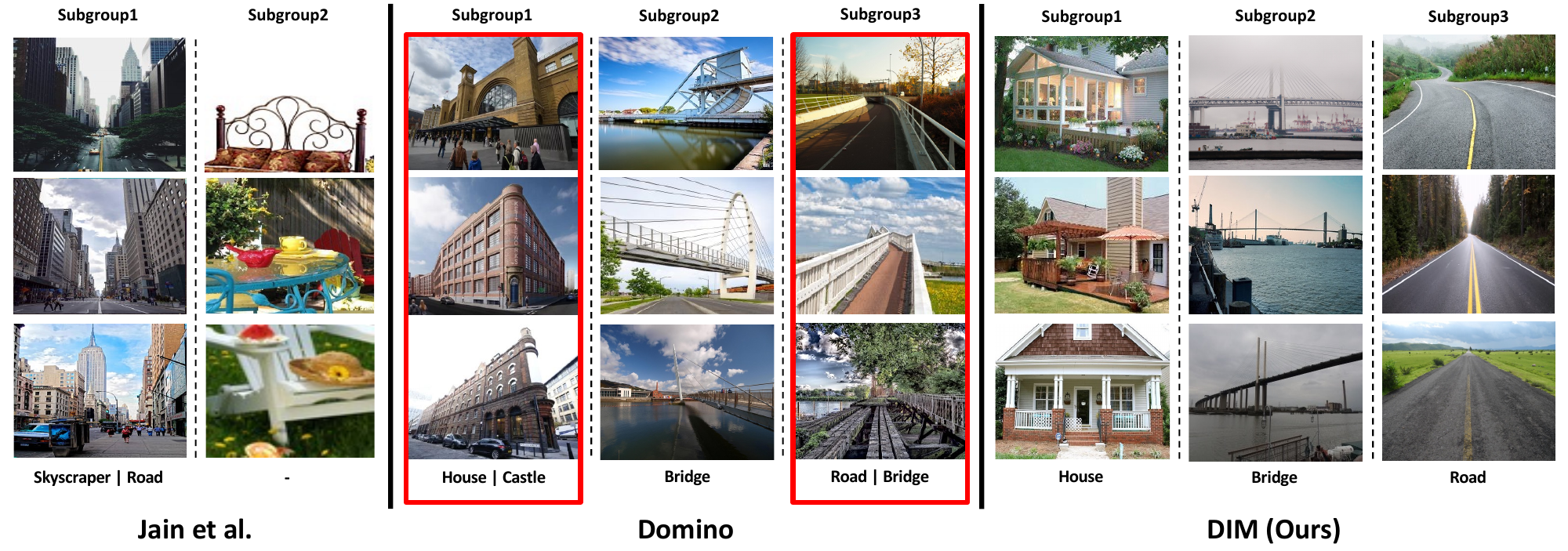}
    \caption{The CLIP-Retrieval results of discovered biased subgroup embeddings in the class ``large man-made outdoor things.'' Images in each column come from the same identified subgroup. \citet{jain2022distilling} inherently discovers two subgroups, positive and negative. Although it successfully detected ``Skyscraper $\mid$ Road,'' it failed to detect the low-performance subgroup. In Domino~\citep{eyuboglu2021domino}, retrieved images from the first subgroup are a mix of ``House'' and ``Castle.''  Similarly, images from the third subgroup confuse the ``Bridge'' and ``Road.''}
    \label{fig:  retrieval_cifar}
    \vspace{-4mm}
\end{figure*}

\begin{table}[t]
    \centering
    \small
    \caption{The overall cosine similarity score $\uparrow$ (averaged over classes) between the subgroups discovered by different methods and the ground-truth subgroups on the CIFAR-100 and Breeds dataset. T.D.: training dynamics; G.T.: ground truths.}
    \label{tab: cosine_sim}
    \resizebox{\linewidth}{!}{
    \begin{tabular}{c|cc|ccc}
\toprule
     Dataset & \citet{jain2022distilling}  & Domino~\citep{eyuboglu2021domino} & \multicolumn{2} {c} {\bfseries \OurMethod{} (Ours)}  & \textcolor{mygray}{\OurMethod-PCA}  \\ \hline
     Supervision & Correctness & Probability & {\bfseries T.D.} & \textcolor{mygray}{G.T.} & \textcolor{mygray}{-}\\
     \hline
     CIFAR-100 &3.06  & 12.20 &{\bfseries 22.97}  &\textcolor{mygray}{22.03} & \textcolor{mygray}{10.76}\\
     Breeds & 3.45   & 10.3 & {\bfseries 24.15} & \textcolor{mygray}{25.26} & \textcolor{mygray}{7.56} \\
\hline
\end{tabular}}
\vspace{-3mm}
\end{table}


\noindent \textbf{Implementation details}. We adopt a consistent approach where models are trained at the class level, and the fine-classes serve as the ground truth for the subgroup discovery task. We train ResNet-18 on CIFAR-100 and ResNet-34 on Breeds from scratch and use ResNet-34 with pre-trained weights for experiments on  Hard ImageNet. All models undergo full training until convergence. We leverage CLIP \citep{radford2021learning} as the foundation model to map images or text to unified representations in the latent feature space. For interpretation, we employ ChatGPT~\citep{ChatGPT} to summarize the concept of grouped description from collected images. More implementation details of our method can be found in \cref{app:imp_det}.

\noindent \textbf{Evaluation metrics.} For the discovery task, we perform a quantitative assessment to provide a robust and transparent comparison. For each class, we assume to identify $G$ subgroup embeddings\footnote{Though the number of ground-truth subgroups is often unknown, we assume it's available for simplicity.} $\{\bm{w}_i\}_1^G$. Then, we compute the representations of the ground-truth subgroups involved in this class via the CLIP text encoder $f_\text{text} \colon \mathcal{T} \rightarrow \mathbb{R}^d$ to ensure semantic interpretability. We adopt the text embedding $\bm{w}^t_i \coloneqq f_\text{text}(\bm{T_\text{prompt}})$ generated from the prompt ``a photo of \{subgroup\}'' to represent the ground-truth subgroup in the same latent space. Building upon this, we enumerate all combinations of subgroup embeddings $\{\bm{w}_i\}_1^G$ and text embeddings $\{\bm{w}^t_i\}_1^G$, aiming to maximize the overall similarity score. This can be framed as the search for a bijective function $\sigma \colon \{1, 2, \ldots, G\} \rightarrow \{1, 2, \ldots, G\}$ that maximizes the sum of matched absolute cosine similarities,
\begin{equation} \label{eq: max_sim_target}
    \setlength\abovedisplayskip{3pt} 
    \setlength\belowdisplayskip{3pt}
    \max_\sigma\sum_{i=1}^{G} |{<\bm{w}^{g}_{i},\bm{w}^{t}_{\sigma(i)}>}|.
\end{equation}

For the mitigation task, we evaluate the classification accuracy of multiple underperforming subgroups (averaged across classes). Classification accuracy on the whole dataset is also provided to show the overall performance. More experimental details can be found in \cref{app:exp_det}.

\subsection{Evaluation on CIFAR-100}\label{sec: exp_cifar}
We begin by validating our framework on the CIFAR-100 dataset with known partitions on the classes. 

\noindent \textbf{Multiple unknown subgroups discovery}. We respectively apply \citet{jain2022distilling}, Domino~\citep{eyuboglu2021domino}, our \OurMethod{}, and \OurMethod-PCA to subgroups discovery for comparative ablation analysis on supervision use. We report the maximum matching similarity between discovered subgroup directions and ground-truth subgroup text embeddings within the latent space in \cref{tab: cosine_sim}. \OurMethod{} (ours) achieves superior performance in terms of \cref{eq: max_sim_target}. Notably, when incorporating training dynamics (T.D.), \OurMethod{} surpass \citet{jain2022distilling} with a significant margin of $10.77$. Furthermore, without supervision, \OurMethod-PCA can only achieve a low similarity of $10.76$, indicating the crucial role of model supervision in the decomposition.

\begin{table}[t]
    \centering
    \small
    \caption{The success rate of bias subgroup detection on CIFAR-100 and Breeds datasets. T.D.: training dynamics; G.T.: ground truths.}
    \label{tab: bias-subgroup-detection_succ_rate}
    \resizebox{\linewidth}{!}{
    \begin{tabular}{c|cc|ccc}
\toprule
     Method & \citet{jain2022distilling}  & Domino~\citep{eyuboglu2021domino} & \multicolumn{2} {c} {\bfseries \OurMethod{} (Ours)}  & \textcolor{mygray}{\OurMethod-PCA}  \\ \hline
     Supervision & Correctness & Probability & {\bfseries T.D.} & \textcolor{mygray}{G.T}.& \textcolor{mygray}{-} \\
     \hline
     CIFAR-100 &45.0  & 35.0 & {\bfseries 57.5}  &\textcolor{mygray}{57.5} & \textcolor{mygray}{52.5}\\
     Breeds & 46.1  & 38.4 & {\bfseries 61.5}  &\textcolor{mygray}{61.5} & \textcolor{mygray}{53.8}\\
\hline
\end{tabular}}
\vspace{-3mm}
\end{table}

\noindent \textbf{Biased subgroup detection}. We evaluate the success rate of the biased subgroup detection on the CIFAR-100 dataset. Specifically, we scrutinize the correspondence between the detected bias subgroups using different methods and the ground-truth bias subgroups. As shown in \cref{tab: bias-subgroup-detection_succ_rate}, \OurMethod{} (ours) accurately identified  $57.5\%$ under-represented subgroups, outperforming Domino~\citep{eyuboglu2021domino}, which achieves a $35.0\%$ success rate. It's worth noting that though \citet{jain2022distilling} achieved a fairly high success rate of $45.0\%$, its similarity between the negative direction and the matched low-performance subgroup is critically low, nearing zero. Quantitatively, the negative direction contributes a mere $0.82$ out of the $3.06$ similarity score achieved by \citet{jain2022distilling} in \cref{tab: cosine_sim}. This indicates that \citet{jain2022distilling} aligns with high-performance subgroups through the identified positive direction, yet its negative direction doesn't effectively represent any biased subgroups. This deficiency further leads to the negative direction not being well interpreted in the next stage.

\begin{table}[t]
\caption{The classification accuracy of ResNet-18 on the CIFAR-100 test set. We present results on the worst two subgroups to study the mitigation performance of multiple biased subgroups. The overall accuracy is also provided for a comprehensive evaluation. }\label{tab: mitigation_cifar}
\resizebox{\columnwidth}{!}{%
\begin{tabular}{ccccl}
\toprule
\multirow{2}{*}{Type} & \multirow{2}{*}{Method}  & \multicolumn{2}{c}{\underline{Worst subgroup accuracy}} & {\multirow{2}{*}{\textcolor{mygray}{Acc.}}} \\
                              &                & 1st   & 2nd   &    \\
                              \hline
-                          & ERM                           & 24.8 & 33.6 & \textcolor{mygray}{44.4}  \\
\hline
\multirow{4}{*}{\begin{tabular}[c]{@{}c@{}}Model-centric\\ (Unsupervised)\end{tabular}} & JTT~\cite{liu2021just}                           & 26.9 & 34.6 & \textcolor{mygray}{48.5} \\
& SubY~\citep{idrissi2022simple}      & 25.1                     & 33.8 & \textcolor{mygray}{45.6}  \\
& LfF~\citep{NEURIPS2020_eddc3427}      & 25.0                     & 33.8 & \textcolor{mygray}{44.3}  \\
& EIIL~\citep{creager2021environment}      & 25.9                    & 34.8 & \textcolor{mygray}{47.2}  \\
                              \hline
\multirow{2}{*}{\begin{tabular}[c]{@{}c@{}}Model-centric\\ (labeled by \citet{jain2022distilling})\end{tabular}}   & gDRO~\citep{sagawa2019distributionally}                           & 26.7 & 34.5 & \textcolor{mygray}{46.9} \\
                              & DI~\citep{wang2020towards}        & 24.3 & 34.2 & \textcolor{mygray}{47.5} \\
\hline
\multirow{2}{*}{\begin{tabular}[c]{@{}c@{}}Model-centric\\ (labeled by Domino~\citep{eyuboglu2021domino})\end{tabular}}   & gDRO~\citep{sagawa2019distributionally}                           & 25.9 & 34.8 & \textcolor{mygray}{47.2} \\
                              & DI~\citep{wang2020towards}        & 25.6 & 35.3 & \textcolor{mygray}{47.1} \\
\hline
\multirow{4}{*}{\begin{tabular}[c]{@{}c@{}}Model-centric\\ (labeled by {\bfseries Ours})\end{tabular}}   & gDRO~\citep{sagawa2019distributionally}                           & 27.2 & 35.8 & \textcolor{mygray}{48.3} \\
                              & Soft-gDRO                      & \textbf{27.2} & \textbf{38.4} & \textcolor{mygray}{\textbf{49.8}} \\
                
                              & DI~\citep{wang2020towards}        & 26.5 & 36.7 & \textcolor{mygray}{48.7} \\
                              & Soft-DI   & 26.8 & 37.1 & \textcolor{mygray}{48.9}\\
\hline
\multirow{2}{*}{Data-centric}  
& \citet{jain2022distilling}     & 33.7 & 41.9  & \textcolor{mygray}{53.1}\\
                              & {\bfseries \OurMethod{} (Ours)}                  & \textbf{35.6} & \textbf{45.1} & \textcolor{mygray}{\textbf{54.7}}\\
\hline
\end{tabular}%
}
\vspace{-3mm}
\end{table}

\begin{figure*}
\centering
    \includegraphics[width=0.95\textwidth]{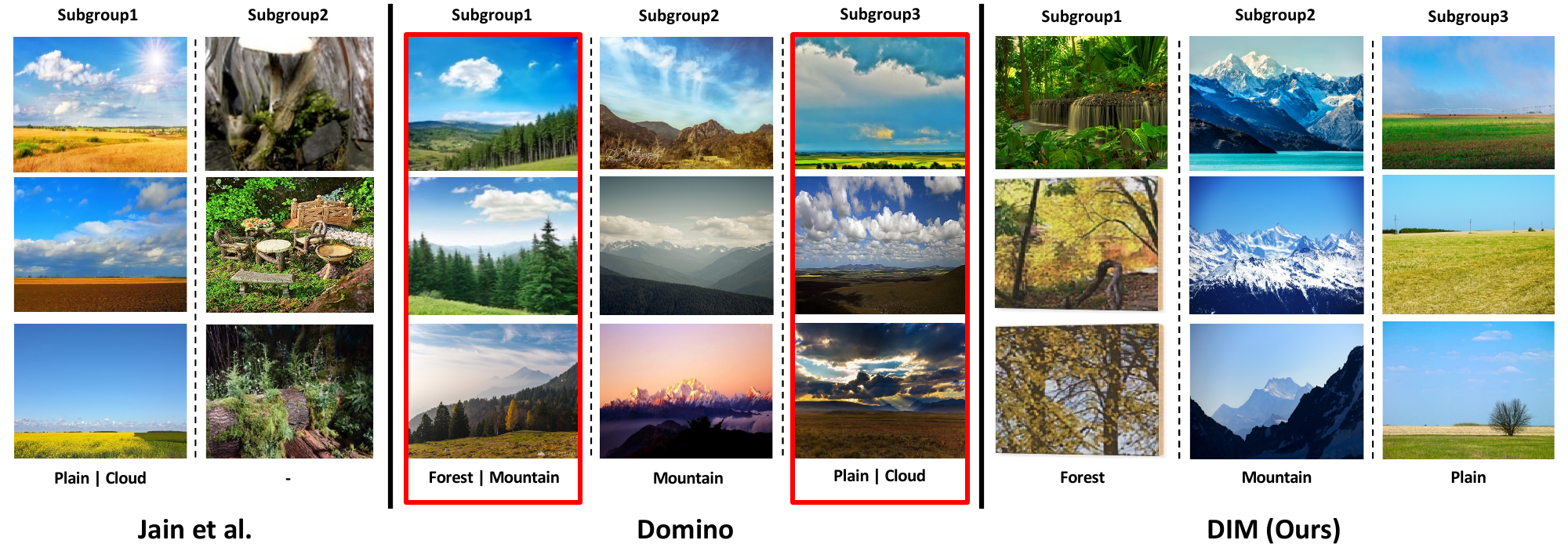}
    \caption{The CLIP-Retrieval results of discovered biased subgroup embeddings in the class ``large natural outdoor scenes.'' Images in each column come from the same identified subgroup. \citet{jain2022distilling} inherently discovers two subgroups, positive and negative. Although it successfully detected ``Plain $\mid$ Cloud,'' it failed to detect the low-performance subgroup. In Domino~\citep{eyuboglu2021domino}, retrieved images from the first subgroup are a mix of ``Forest'' and ``Mountain.''  Similarly, images from the third subgroup confuse the ``Plain'' and ``Cloud.''}
    \label{fig:  retrieval_cifar2}
    \vspace{-4mm}
\end{figure*}

\noindent \textbf{Subgroup interpretation}. Both our method and \citet{jain2022distilling} capitalize on the latent space of CLIP, allowing us to decode discovered subgroups to images using CLIP-Retrieval naturally. 
We present the retrieval outcomes for the discovered subgroups within the class ``large man-made outdoor things'' using \citet{jain2022distilling}, Domino~\citep{eyuboglu2021domino}, and our \OurMethod{} in \cref{fig:  retrieval_cifar}. As displayed, \citet{jain2022distilling} was partial to the well-performance group but failed to illustrate the biased one. Due to space constraints, we selectively exhibit three subgroups for Domino~\citep{eyuboglu2021domino} and our \OurMethod{}. We can see that \OurMethod{} distinctly and accurately embodied two low-performance subgroups, ``bridge'' and ``road.'' In contrast, though Domino~\citep{eyuboglu2021domino} discovered multiple unknown subgroups, it confused multiple concepts. This finding corroborates our argument that using proxies induces misinterpretation. More results on CIFAR-100 can be found in \cref{app:more_detail_cifar}.

\noindent \textbf{Bias mitigation}. We apply the model- and data-centric strategies to bias mitigation. For model-centric methods (groupDRO~\citep{sagawa2019distributionally} and DI~\citep{wang2020towards}), we use the hard label generated from \citet{jain2022distilling} and Domino~\citep{eyuboglu2021domino} for supervised training. Our method provides both hard and soft labels for mitigation. For the data-centric strategy, we select the top $20\%$ of images with the highest scores on the two bias subgroups identified by \citet{jain2022distilling} and our \OurMethod{}. The results are compiled in \cref{tab: mitigation_cifar}. Compared with unsupervised mitigation methods, discovered subgroups in supervised methods boost the mitigation performance with a clear gap of $2.2\%$ on average. Our \OurMethod{} provides a more precise subgroup discovery for mitigation, achieving a better mitigation performance improvement with up to $1.8\%$ compared with \citet{jain2022distilling} and Domino~\citep{eyuboglu2021domino}. By filtering multiple under-represented subgroups, \OurMethod{} has a significant improvement of $3.2\%$ over \citet{jain2022distilling}, which not only filters a single subgroup but also falls short in precisely representing it. The statistical significance  is discussed in \cref{appendix:ss}.

\begin{figure*}[t]
    \centering
    \includegraphics[width=0.95\linewidth]{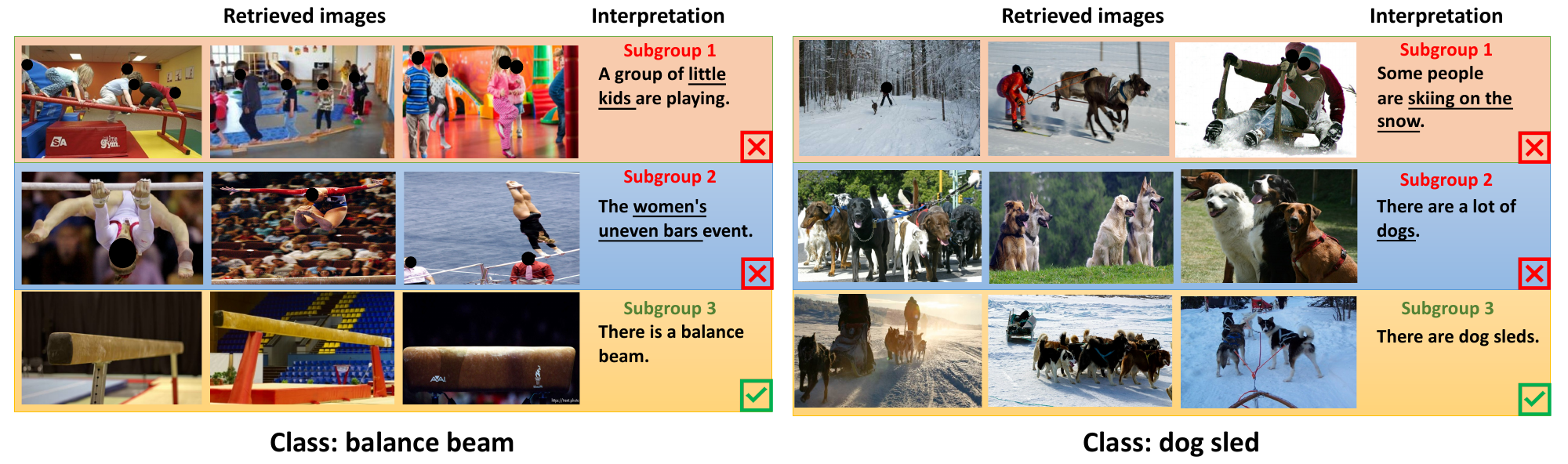}
    \caption{Example of subgroup interpretation on Hard ImageNet. The first two rows are the retrieval images of identified biased subgroups and corresponding summary descriptions by ChatGPT based on metadata. The last row is from the high-performance subgroup.}
    \label{fig: hard_imagenet_exp}
    \vspace{-4mm}
\end{figure*}

\begin{table}[t]
\caption{The classification accuracy of ResNet-34 on the Breeds test set. We present results on the worst four subgroups to study the mitigation performance of multiple biased subgroups. The overall accuracy is also provided for a comprehensive evaluation. }\label{tab: mitigation_breeds}
\resizebox{\columnwidth}{!}{%
\begin{tabular}{cccccccl}
\toprule
\multirow{2}{*}{Type} & \multirow{2}{*}{Method}  & \multicolumn{5}{c}{\underline{Worst subgroup accuracy}}& {\multirow{2}{*}{\textcolor{mygray}{Acc.}}} \\
                              &               & 1st   & 2nd   & 3rd   & 4th & 5th&   \\\hline
-                          & ERM                       & 52.4 & 58.4 & 64.4 & 69.0& 71.4& \textcolor{mygray}{72.7} \\\hline
\multirow{4}{*}{\begin{tabular}[c]{@{}c@{}}Model-centric\\ (Unsupervised)\end{tabular}} & JTT~\citep{liu2021just} & 64.7 & 70.0 & 73.2 & 75.9&  79.2 &\textcolor{mygray}{80.0} \\
& SubY~\citep{idrissi2022simple}      & 60.8                     & 62.9 & 67.5 & 70.3& 73.5& \textcolor{mygray}{74.1} \\
& LfF~\citep{NEURIPS2020_eddc3427}      &   63.3                   & 64.1 & 69.5 & 72.1& 76.9& \textcolor{mygray}{76.9} \\
& EIIL~\citep{creager2021environment}      & 62.8                     & 64.5 & 70.6 & 73.2& 75.3& \textcolor{mygray}{77.0} \\
                              \hline
\multirow{2}{*}{\begin{tabular}[c]{@{}c@{}}Model-centric\\ (labeled by \citet{jain2022distilling})\end{tabular}}   & gDRO~\citep{sagawa2019distributionally}                           & 63.2 & 69.6 & 72.8 & 74.1 &  77.9&\textcolor{mygray}{76.9}\\
                              & DI~\citep{wang2020towards}        & 65.9 & 71.5 & 76.2 & 83.6& 85.2& \textcolor{mygray}{84.4}  \\\hline
\multirow{2}{*}{\begin{tabular}[c]{@{}c@{}}Model-centric\\ (labeled by Domino~\citep{eyuboglu2021domino})\end{tabular}}   & gDRO~\citep{sagawa2019distributionally}                           & 65.2 & 71.3 & 74.2 & 76.5 &  78.2 &\textcolor{mygray}{78.3}\\
                              & DI~\citep{wang2020towards}         & 66.1 & 72.3 & 77.9 & 84.2& 85.3& \textcolor{mygray}{85.0}  \\\hline
\multirow{4}{*}{\begin{tabular}[c]{@{}c@{}}Model-centric\\ (labeled by {\bfseries Ours})\end{tabular}}   & gDRO~\citep{sagawa2019distributionally}                           & 66.8 & 72.5 & 75.7 & 78.2 &  82.5&\textcolor{mygray}{80.7}\\
                              & Soft-gDRO                     & \textbf{69.4} & \textbf{75.3} & 79.7 & 83.0& 84.1& \textcolor{mygray}{82.5}  \\
                              & DI~\citep{wang2020towards}        & 67.3 & 74.2 & 79.8 & 87.6& 88.5& \textcolor{mygray}{86.0} \\
                              & Soft-DI   & 68.2 & \textbf{75.3} & \textbf{81.4} & \textbf{88.3}& \textbf{89.6}& \textcolor{mygray}{\textbf{86.7}}\\\hline
\multirow{2}{*}{Data-centric}   
                                & \citet{jain2022distilling}             &71.6            & 79.5 & 82.9 &  89.3& 89.9 & \textcolor{mygray}{89.7}\\
                              & {\bfseries \OurMethod{} (Ours)}   & \textbf{72.5} & \textbf{82.1} & \textbf{84.6} & \textbf{91.3}& \textbf{91.5} & \textcolor{mygray}{\textbf{89.7}}\\
\hline
\end{tabular}%
}\vspace{-4mm}
\end{table}

\subsection{Evaluation on Breeds}
Our evaluation extends to Breeds~\citep{santurkar2021breeds} dataset, where we test multiple unknown subgroup discovery and bias mitigation. Breeds, as outlined in \cref{sec: setup}, mount a more serious challenge in scalability with its fine-grained partition into $130$ ground-truth subgroups across $13$ classes.

We begin by discovering multiple unknown subgroups. As shown in \cref{tab: cosine_sim}, compared to baselines, our \OurMethod{} achieves the highest similarity score with a significant lead of $13.85$ for the discovery of unknown subgroups.

Subsequently, we leverage identified subgroups to label data in preparation for different model-centric mitigation methods. We report the classification accuracy of the worst $4$ subgroup to evaluate the mitigation performance. The results, as depicted in \cref{tab: mitigation_breeds}, suggest that our proposed \OurMethod{} provides accurate subgroup supervision for mitigation methods, improving the worst subgroup accuracy of $17.3\%$ on average compared with the baseline (ERM). This constitutes an improvement margin of $4.33\%$ versus other subgroup discovery methods, \citet{jain2022distilling} and Domino~\citep{eyuboglu2021domino}. More results on Breeds can be found in \cref{app:more_detail_breeds}.


\subsection{Interpreting Hard ImageNet}
Unlike CIFAR-100 and Breeds, which have known well-defined class partitions into different subgroups, Hard ImageNet hasn't been extensively analyzed for the existence of multiple biases, presenting unique challenges. Previous studies primarily focus on using counterfactual images to explain the failure modes of models from the single-image level but lack analysis on the subgroup level. We apply our method to discover implicit subgroups within Hard ImageNet.

Following the same setting in \citep{moayeri2022hard}, we study the ResNet-50 model first pre-trained on the full ImageNet dataset and then fine-tuned on the Hard ImageNet dataset. In our \OurMethod{} framework, We set the number of subgroups to be discovered in each class to $5$ for latent space decomposition. The ablation accuracy is used to evaluate the model's performance on Hard ImageNet, which is the performance drop introduced by masking the target object. A higher ablation accuracy means that the model relies more on spurious features, indicating the existence of implicit bias. We use the ablation accuracy of the validation set to identify biased subgroups.

Using our proposed \OurMethod, we discover $5$ subgroups for each class. To explain the failure modes of the studied model, we retrieve images with descriptions (metadata) belonging to the worst $2$ subgroups. We then use ChatGPT to summarize the aggregated descriptions into coherent concepts. For comparison, we also analyze images from the best-performing subgroup. We present examples of the class ``balance beam'' and ``dog sled'' in \cref{fig: hard_imagenet_exp}. Our findings reveal that the model is vulnerable to spurious correlations, \textit{i.e.}, the ``population'' and ``horizontal bar'' in the class ``balance beam.'' In the class ``dog sled,'' our retrieval images and interpretation show that the model may make bad decisions based on necessary but insufficient objects like ``snow'' and ``dogs.'' These results surface the failure modes, \textit{i.e.}, the presence of multiple spurious features, significantly impairing model's robustness. More results on Hard ImageNet can be found in \cref{app:more_detail_hard}.

\begin{table}[t]
\caption{Ablation study on the $n$, number of components, in discovering the unknown subgroups. We report different methods' maximum matching similarity $\uparrow$(\%).}\label{tab: aba_n}
\resizebox{\columnwidth}{!}{%
\begin{tabular}{c|ccccccccc}
\toprule
     \# of comp. ($n$) & 2 & 3 & 4 & 5 & 6 & 7 & 8  &  9 & 10 \\
     \hline
     \citet{jain2022distilling}& 3.1& \multicolumn{8}{c}{--------------------------------------------------------------} \\
     Domino~\citep{eyuboglu2021domino}& 6.1& 8.4 &10.4 & 12.2& 12.7& 12.9& 13.3&13.4 & 13.5 \\
     \bfseries \OurMethod{} (Ours)& \bf{9.9} & \bf{14.9} & \bf{19.4}& \bf{22.9}& \bf{25.2}& \bf{26.6}& \bf{27.8}&\bf{28.7} & \bf{29.9}\\
     \hline
\end{tabular}
}\vspace{-4mm}
\end{table}

\subsection{Ablation study}
\noindent \textbf{On the number of subgroups to be discovered}. On the CIFAR-100 dataset, we vary the number $n$ of iteration, corresponding to the subgroups for discovery, from $2$ to $10$, match the discovered embeddings to the ground-truth subgroups using the same strategy as \cref{eq: max_sim_target}, and report the maximum similarity score. As displayed in \cref{tab: aba_n}, while \citet{jain2022distilling} fails to identify multiple subgroups and Domino~\citep{eyuboglu2021domino} lacks the capacity of fine-grained partitions due to the dataset-level clustering, our \OurMethod{} demonstrates a better performance when varying $n$, showing great stability and scalability in the hyperparameter selection by using dimension reduction.

\noindent \textbf{On the use of supervision}. In our \OurMethod, the model supervision is managed to guide the latent space decomposition. Without supervision, the decomposition stage may fail to reflect the subgroups learned by the model. To support our argument, we remove supervision at the decomposition stage and use \OurMethod-PCA to discover unknown subgroup directions. The performance degradation in numerical results ($14.4\%\downarrow$ for subgroup discovery in \cref{tab: cosine_sim} and $6.25\%\downarrow$ for biased subgroup detection in \cref{tab: bias-subgroup-detection_succ_rate} on average) sufficiently demonstrates the crucial role of supervision. Supportive retrieval results can be found in \cref{app:cifar_result}.


\section{Conclusion}
In this work, we present a novel approach to the problem of multiple unknown biased subgroups in image classifiers. The proposed \OurMethod{} includes three stages: decomposition,  interpretation, and mitigation. We employ model supervision to guide the latent space decomposition and reveal distinct subgroup directions. Then, we identify the biased subgroups and interpret the model failures. The discovered subgroups can be further integrated into the downstream mitigation stage. Our methodology, validated through experiments on CIFAR-100, Breeds, and notably Hard ImageNet, effectively detects subgroups and uncovers new model failure modes related to spurious correlations. The use of model training dynamics as supervision is pivotal in this process, yet selecting optimal supervisory signals for enhanced subgroup representation remains an open area for future research. This study contributes to the field by providing a methodological framework for understanding and improving image classifiers' subgroup robustness.

{
    \small
    \bibliographystyle{ieeenatetal_fullname}
    \bibliography{main}
}

\clearpage
\onecolumn
\appendix
\renewcommand{\appendixpagename}{Appendix}
\appendixpage

\section{Implementation Details of DIM}
\label{app:imp_det}

\subsection{Preliminaries}
In a standard image classification setting with subgroup information, inputs $\bm{x}$ in image space $\mathcal{X}$, labels $y$ in class space $\mathcal{Y}$, and subgroups $g$ in subgroup space $\mathcal{G}$ follows some certain distribution $P$. Specifically, training data, validation data, and test data are respectively drawn from distributions $\hat{P}_{train}$, $\hat{P}_{val}$, and $\hat{P}_{test}$. An image classifier trained on training data $(\bm{x}, y, g)$, where $g$ is not available to the classifier, can be denoted as a function $\phi_\theta \colon \mathcal{X} \rightarrow \mathcal{Y}$ that predict the class of a given image. Assume there are $L$ annotated classes in the dataset, making $\mathcal{Y}=\{1,2,..., L\}$. For each class, we hypothesize the inputs can be further divided into $G$ subgroups. The subgroup is defined as a smaller, more specific category within a larger class, representing a partition based on certain characteristics or features, like the ``dolphin'' and ``beaver'' in the class ``aquatic mammals''. 

In this work, for simplicity, we hypothesize that $G \in \mathbb{Z}^+$ is a constant value across classes. It's also supported by the fact that while data can be partitioned into subgroups in various ways through different standards, the partitioning that affects the model's output most is what we are most interested in. Due to under-representation or other difficulties, the image classifier exhibits low performance on certain subgroups within one class. We define bias subgroups as subgroups with lower classification accuracy than the median, which implies $k=\lfloor \frac{G}{2} \rfloor$ biased subgroups. The primary objective of this work is to discover multiple bias subgroups ($k \geq 2$), thereby implying $\lfloor \frac{G}{2} \rfloor \geq 2$ and then $G \geq 4$.

Either mean or median is reasonable in this context. We choose the median here for simplicity. If using mean, a constant number of subgroups $G$ across classes does not guarantee a constant number of biased subgroups due to different cases of subgroup classification accuracy. For example, accuracies of $(0.1,0.2,0.8,0.9)$ lead to $2$ subgroups lower than the mean. However, accuracies like $(0.1,0.7,0.8,0.9)$ results in only one bias subgroup.

\subsection{Decomposition}
In the decomposition stage of \OurMethod{}, we apply the partial least squares (PLS) method to discover the embeddings of multiple unknown subgroups. Here, we give a detailed overview of the PLS.

\noindent \textbf{PLS Details.} We consider the input $\bm{x}$ and the response $\bm{z}$. PLS consists of the following steps iteratively repeated n times (for n components):
\begin{enumerate}
\item searching for paired directions that maximize covariance between the corresponding components in the input (observation) space and response (supervision) space.
\begin{equation}
    \setlength\abovedisplayskip{3pt} 
    \setlength\belowdisplayskip{3pt}
    \begin{aligned} \label{appeq: PLS}
    &\max_{\bm{w}_{i},~\bm{h}_{i}} \mathbb{E}_{\hat{P}_l}(u_i v_i)=\max_{\bm{w}_{i},~\bm{h}_{i}} \mathbb{E}_{\hat{P}_l}\left((\bm{w}_{i}^T \widehat{\bm{x}}_i)(\bm{h}_{i}^T \bm{z}_{\bm{x},i})\right) \\
    &~~\textit{s.t.}~ \Vert \bm{w}_{i} \Vert=1,~\Vert \bm{h}_{i} \Vert=1
    \end{aligned}
\end{equation}
where $\bm{w}_{i} \in \mathbb{R}^d$, $\bm{h}_{i} \in\mathbb{R}^M$ are the discovered directions and $u_i \coloneqq \bm{w}_{i}^T \widehat{\bm{x}}_i$, $v_i \coloneqq \bm{h}_{i}^T \bm{z}_{\bm{x}, i}$ are the corresponding components (also called input score and response score). In the matrix notation, where the input $X \in \mathbb{R}^{N\times D}$ and the response $Z \in \mathbb{R}^{N\times M}$ are matrices ($N$ is the number of samples), $\bm{w}_{i}$ and $\bm{h}_{i}$ are the first left and right singular vectors of the cross-covariance matrix $X^TZ$.
\item performing least squares regression on the scores for input and response.
\begin{equation}
    \setlength\abovedisplayskip{3pt} 
    \setlength\belowdisplayskip{3pt}
    \left\{
         \begin{array}{lr}
         \widehat{\bm{x}}_i=u_i\bm{\alpha}_i+\bm{\epsilon}_i\\
         \bm{z}_{\bm{x},i}=v_i\bm{\beta}_i+\bm{\eta}_i,
         \end{array}
    \right.
\end{equation}
where $\bm{\alpha}_i$, $\bm{\beta}_i$ are regression coefficients and $\bm{\epsilon}_i$, $\bm{\eta}_i$ are the remainders.
\item deflating the inputs and responses by subtracting the approximation modeled by the regression. $\widehat{\bm{x}}_{i+1} \coloneqq \widehat{\bm{x}}_i -u_i\bm{\alpha}_i=\bm{\epsilon}_i$ and $\bm{z}_{\bm{x},i+1} \coloneqq \bm{z}_{\bm{x},i} -v_i\bm{\beta}_i=\bm{\eta}_i$.
\end{enumerate}

In our problem, we use the CLIP embedding of images as the $\bm{x}$ and the model supervision as $\bm{z}$. With such a design, we can derive the principal components of image features mostly aligned with the changes of the supervision, achieving the supervised decomposition.

\subsection{Interpretation}
To exploit existing coarse-grained knowledge, we employ text embeddings generated from class-specific text prompts, ``a photo of \{class\},'' as retrieval bases. For instance, within the ``large man-made outdoor things'' class, we first compute the text embedding from ``a photo of large man-made outdoor things.'' We then add each discovered subgroup embedding to it and proceed to retrieve images and corresponding metadata using the resultant normalized embeddings.

\section{Bias Mitigation with Soft-label Strategy}
\label{app:alg_soft}
In our paper, we have provided the model-centric strategy with Soft-DI in \cref{sec: method}. The soft-label strategy can also be applied to  the groupDRO method as follows,

\noindent \textbf{\textit{Case study of model-centric strategy with soft gDRO.}} \quad Consider the empirical distribution on the training data $\hat{P}$. For some assignment of weights $\bm{q}=(q_1,...,q_N)\in \Delta_N$, where $\Delta_N$ is the $(m-1)$-dimensional probability simplex, the expected loss and the way to update  $\bm{q}$ in original groupDRO~\citep{sagawa2019distributionally} is:
\begin{align}
\setlength\abovedisplayskip{3pt} 
    \setlength\belowdisplayskip{3pt}
    \mathcal {L} &= \mathbb{E}_{(\bm{x},y,g) \sim \hat{P}} \left( q_g \ell(\theta; (\bm{x},y)) \right) \\
    \bm{q}^{(t)}_n &= \bm{q}^{(t-1)}_n \exp\left(\eta_q \mathbb{E}_{\hat{P}}(\ell(\theta; (\bm{x},y))|g=n)\right)
\end{align}
where $g \in \{1,2,...,N\}$ is the hard group label of data.

We define soft-label version groupDRO~\citep{sagawa2019distributionally} expected loss and the way to update weights $\bm{q}$ as following:
\begin{align}
\setlength\abovedisplayskip{3pt} 
    \setlength\belowdisplayskip{3pt}
    \mathcal {L} &= \mathbb{E}_{(\bm{x},y,\bm{g}) \sim \hat{P}} \left(\bm{q}^T\bm{g} \ell(\theta; (\bm{x},y))\right)\\
    \bm{q}^{(t)}_n &= \bm{q}^{(t-1)}_n \exp\left(\eta_q \mathbb{E}_{\hat{P}}(g_n \ell(\theta; (\bm{x},y)))\right)
\end{align}
where $\bm{g} = (g_1,...,g_N) \in \Delta_N$ is the soft label of data.

\section{Experiment Details}
\label{app:exp_det}

\subsection{The Selection of Supervision}
\label{app:selet_sup}
In our experiments, we use three kinds of supervision: correctness, logit, and loss. All are in the form of training dynamics. For each image $\bm{x}$ and the corresponding label $y$, we define correctness as whether the image classifier $\phi_\theta$ correctly outputs the label, $\mathbbm{1}\{\phi_\theta(\bm{x})=y\}$. The logit is the unnormalized (without soft-max activation) final score of the image classifier. For loss, we adopt the cross-entropy loss. Specifically, we train the image classifier from scratch and record the correctness, logit, and loss for each image in each epoch. Then, we concatenate all this information together to supervise the decomposition of image features. For training of $t$ epochs, the final corresponding supervision $z$ of each image is a vector of length $3t$.

The use of supervision may have a different impact on the decomposition of image features. The logit provides information on image features that the biased model learns. Correctness helps the decomposition align with the direction that affects the image classifier's performance. Loss is a fine-grained correctness. 

\subsection{Bias Mitigation}
\begin{wraptable}{r}{0.5\textwidth}
\caption{The classification accuracy (\textbf{including the error bar}) of ResNet-18 on the CIFAR-100 test set. }\label{tab: aba_mitigation_cifar}
\resizebox{0.5\textwidth}{!}{%
\begin{tabular}{ccccl}
\toprule
\multirow{2}{*}{Type} & \multirow{2}{*}{Method}  & \multicolumn{2}{c}{\underline{Worst subgroup accuracy}} & {\multirow{2}{*}{\textcolor{mygray}{Acc.}}} \\
                              &                & 1st   & 2nd   &    \\
                              \hline
-                          & ERM                           & 24.8$_{\pm 0.09}$ & 33.6$_{\pm 0.12}$ & \textcolor{mygray}{44.4$_{\pm 0.11}$}  \\
\hline
\multirow{4}{*}{\begin{tabular}[c]{@{}c@{}}Model-centric\\ (Unsupervised)\end{tabular}} & JTT~\cite{liu2021just}                           & 26.9$_{\pm 0.24}$ & 34.6$_{\pm 0.30}$ & \textcolor{mygray}{48.5$_{\pm 0.25}$} \\
& SubY~\citep{idrissi2022simple}      & 25.1$_{\pm 0.16}$                     & 33.8$_{\pm 0.16}$ & \textcolor{mygray}{45.6$_{\pm 0.13}$}  \\
& LfF~\citep{NEURIPS2020_eddc3427}      & 25.0$_{\pm 0.22}$                     & 33.8$_{\pm 0.18}$ & \textcolor{mygray}{44.3$_{\pm 0.15}$}  \\
& EIIL~\citep{creager2021environment}      & 25.9$_{\pm 0.09}$                    & 34.8$_{\pm 0.07}$ & \textcolor{mygray}{47.2$_{\pm 0.10}$}  \\
                              \hline
\multirow{2}{*}{\begin{tabular}[c]{@{}c@{}}Model-centric\\ (labeled by \citet{jain2022distilling})\end{tabular}}   & gDRO~\citep{sagawa2019distributionally}                           & 26.7$_{\pm 0.11}$ & 34.5$_{\pm 0.15}$ & \textcolor{mygray}{46.9$_{\pm 0.12}$} \\
                              & DI~\citep{wang2020towards}        & 24.3$_{\pm 0.06}$ & 34.2$_{\pm 0.14}$ & \textcolor{mygray}{47.5$_{\pm 0.10}$} \\
\hline
\multirow{2}{*}{\begin{tabular}[c]{@{}c@{}}Model-centric\\ (labeled by Domino~\citep{eyuboglu2021domino})\end{tabular}}   & gDRO~\citep{sagawa2019distributionally}                           & 25.9$_{\pm 0.09}$ & 34.8$_{\pm 0.07}$ & \textcolor{mygray}{47.2$_{\pm 0.07}$} \\
                              & DI~\citep{wang2020towards}        & 25.6$_{\pm 0.18}$ & 35.3$_{\pm 0.22}$ & \textcolor{mygray}{47.1$_{\pm 0.19}$} \\
\hline
\multirow{4}{*}{\begin{tabular}[c]{@{}c@{}}Model-centric\\ (labeled by {\bfseries Ours})\end{tabular}}   & gDRO~\citep{sagawa2019distributionally}                           & 27.2$_{\pm 0.04}$ & 35.8$_{\pm 0.07}$ & \textcolor{mygray}{48.3$_{\pm 0.06}$} \\
                              & Soft-gDRO                      & \textbf{27.2$_{\pm 0.08}$} & \textbf{38.4$_{\pm 0.12}$} & \textcolor{mygray}{\textbf{49.8$_{\pm 0.09}$}} \\
                
                              & DI~\citep{wang2020towards}        & 26.5$_{\pm 0.05}$ & 36.7$_{\pm 0.07}$ & \textcolor{mygray}{48.7$_{\pm 0.08}$} \\
                              & Soft-DI   & 26.8$_{\pm 0.06}$ & 37.1$_{\pm 0.04}$ & \textcolor{mygray}{48.9$_{\pm 0.07}$}\\
\hline
\multirow{2}{*}{Data-centric}  
& \citet{jain2022distilling}     & 33.7$_{\pm 0.15}$ & 41.9$_{\pm 0.07}$  & \textcolor{mygray}{53.1$_{\pm 0.12}$}\\
                              & {\bfseries \OurMethod{} (Ours)}                  & \textbf{35.6$_{\pm 0.10}$} & \textbf{45.1$_{\pm 0.04}$} & \textcolor{mygray}{\textbf{54.7$_{\pm 0.11}$}}\\
\hline
\end{tabular}%
}
\end{wraptable}
In our evaluation, we adopt two types of methods to mitigate the biased behavior of the model, namely the unsupervised and supervised methods. For the hyper-parameter settings, the details are presented as follows:
\begin{enumerate}
    \item For JTT~\citep{liu2021just},  by grid-search on the hyper-parameters, we set the number of epochs for first-time training $T$ as $10$, the up-sampling factor $\lambda_{up}$ as $\frac{\left|\text{Training set}\right|}{\left|\text{Error set}\right|}$. We set the total number of training epochs as the same as the vanilla training. 
    \item For SubY~\citep{idrissi2022simple}, no hyper-parameter is required. 
    \item For LfF~\citep{NEURIPS2020_eddc3427}, we tune the hyper-parameter $q$ by grid searching over $q \in \{0.1, 0.3, 0.5, 0.7, 0.9\}$.
    \item For EIIL~\citep{creager2021environment}, no hyperparameter is required. 
    \item For gDRO~\citep{sagawa2019distributionally} , we set the group number as the number of subgroups in each class, \textit{i.e.}, $5$in CIFAR-100, $13$ in Breeds, and set $\eta$ as $0.1$. 
    \item For DI,  we set the group number as the number of subgroups in each class, \textit{i.e.}, $5$ in CIFAR-100, and $13$ in Breeds.
\end{enumerate}

For all of the baseline methods, we use the Adam optimizer to train the model. The grid search has fine-tuned hyperparameters to achieve the best performance. 

\noindent \textbf{Statistical significance}\label{appendix:ss} As shown in \cref{tab: aba_mitigation_cifar}, we train each model for $5$ times and report the mean accuracy for a fair comparison. Specifically, for the most efficient data-centric mitigation of ResNet-18 on CIFAR-10, our method has an average accuracy on the worst two subgroups of $40.35\pm 0.07 \%$, while that of the runner-up method~\citep{jain2022distilling} is $37.80\pm0.11\%$.The performance gap without overlap indicates significant performance improvement.

\begin{figure}
    \centering
    \includegraphics[width=\linewidth]{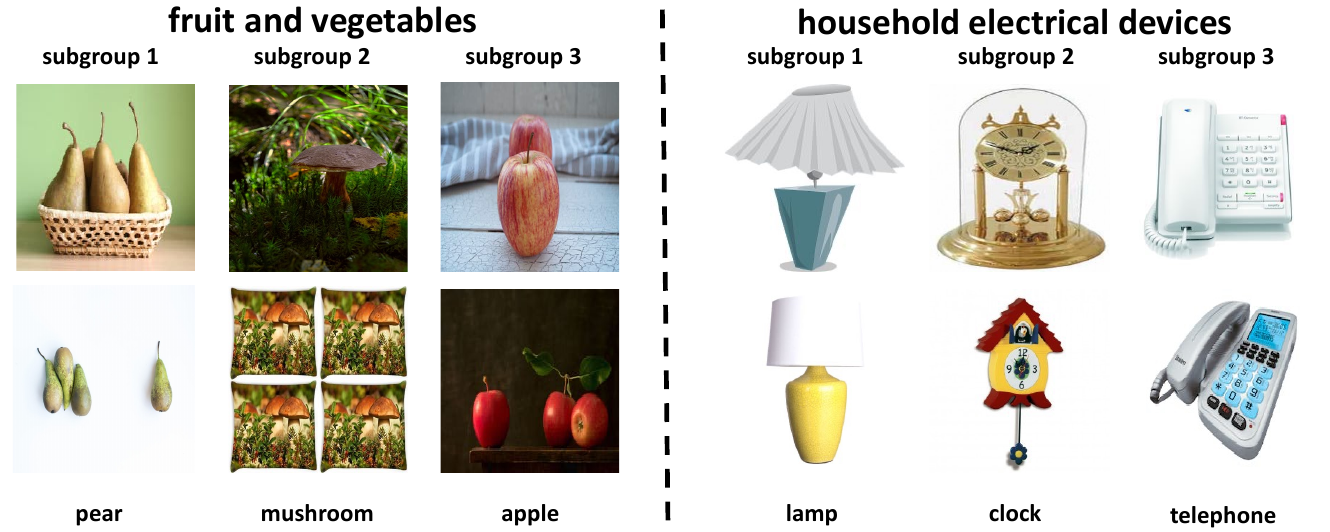}
    \caption{The CLIP Retrieval results of discovered subgroups in the class of ``fruit and vegetables'' and ``household electrical devices'' respectively. Images in each column come from the same identified subgroup.}
    \label{appfig:cifar100_1}
\end{figure}

\begin{figure}
    \centering
    \includegraphics[width=\linewidth]{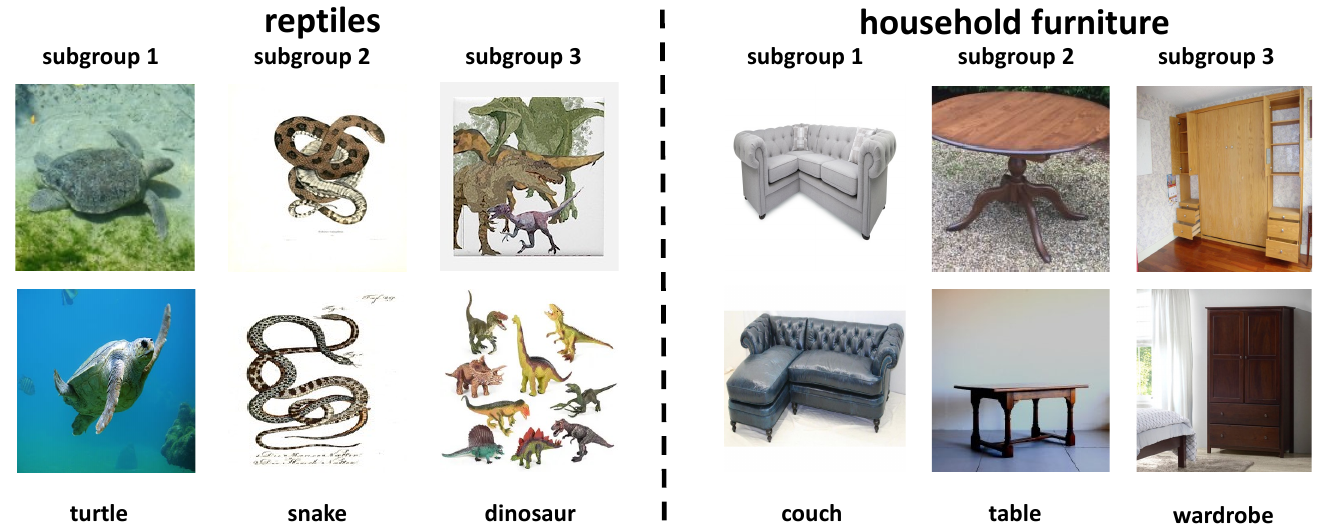}
    \caption{The CLIP Retrieval results of discovered subgroups in the class of ``reptiles'' and ``household furniture
'' respectively. Images in each column come from the same identified subgroup.}
    \label{appfig:cifar100_2}
\end{figure}

\section{Experiments on CIFAR-100}
\label{app:more_detail_cifar}
\subsection{Experimental Details}
We train ResNet-18 on the CIFAR-100 dataset from scratch. We use SGD as the optimizer. We set the learning rate as $0.1$ as the learning rate and the batch size as $128$.  We set the number of training epochs as $50$ and used early stop with validation loss as the criterion to eliminate the overfitting problem. 

\subsection{Subgroup Interpretation}
\label{app:cifar_result}

We show examples of detecting multiple subgroups within the CIFAR-100 dataset. As displayed in \cref{appfig:cifar100_1}, our method can accurately capture multiple unknown subgroups, the ``pear'',  ``mushroom'', and ``apple'' in the ``fruit and vegetables'' class, ``lamp'', ``clock'', and ``telephone'' in the ``household electrical devices'' class. ``pear'',  ``mushroom'', ``lamp'', and ``clock'' are low-performance subgroups. ``apple'' and ``telephone'' are easy subgroups. In \cref{appfig:cifar100_2}, our \OurMethod{} successfully identified subgroups, ``turtle'', ``snake'', and ``dinosaur'', in the class ``reptiles'' and ``household furniture'', and ``couch'', ``table'', and ``wardrobe'' in the class ``household funiture''. ``turtle'',  ``snake'', ``couch'', and ``table'' are low-performance subgroups. ``dinosaur'' and ``wardrobe'' are easy subgroups.

\begin{wrapfigure}[14]{r}{0.3\textwidth}
\centering
\includegraphics[width=\linewidth]{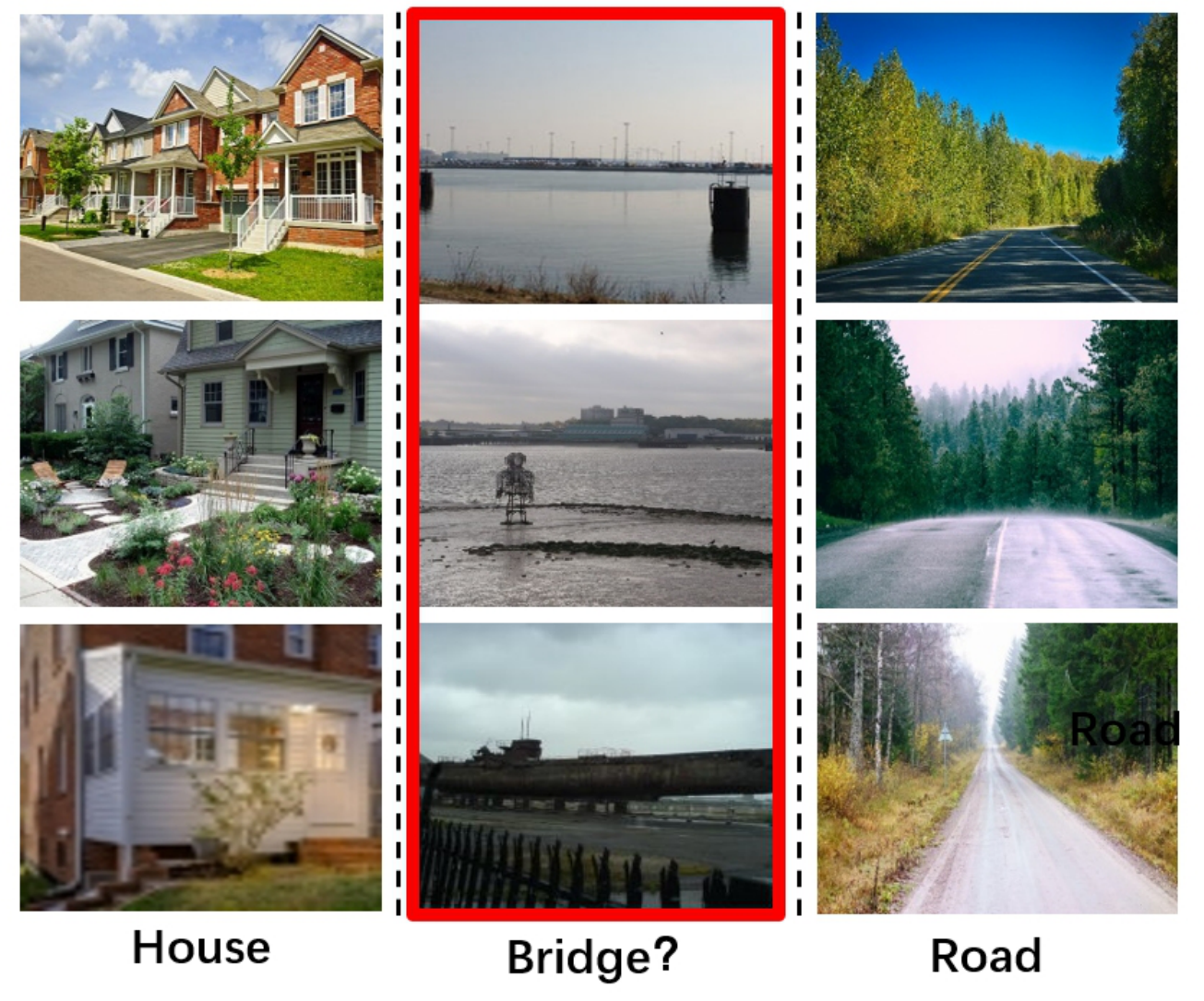}
\caption{The CLIP-Retrieval results for interpreting the multiple subgroups within the ``large man-made outdoor things'' class discovered by \OurMethod-PCA on the CIFAR-100 dataset.} 
\label{fig: pca_results}
\end{wrapfigure}

\noindent \textbf{Ablation study on supervision}. We conduct experiments on studying the impact of supervision in the decomposition of our proposed \OurMethod{}. Without supervision, the use of the PLS at the decomposition stage degrades to the PCA. We replace the PLS with PCA in \OurMethod{}. As shown in \cref{fig: pca_results}, in the ``large man-made outdoor things'' class of CIFAR-100, we can see that the PCA can not accurately discover the ``bridge'' subgroup, which is confused by the spurious correlation between bridge and water. For comparison in \cref{fig:  retrieval_cifar}, the PLS can accurately discover the ``bridge'' subgroup. 

It supports our argument that it is impossible to discover the multiple subgroups in the image classifier without the supervision of the studied model.

\subsection{Bias Mitigation}
\noindent \textbf{Ablation study on number of subgroups}. We conduct experiments to study the impact of the number of subgroups to be discovered, "$n$", on the classification accuracy in data-centric mitigation. When we reduce $n$ from $5$ to $3$ on the CIFAR-100 dataset, the mean accuracy of the worst two subgroups is reduced from ${40.35\%}$ to ${39.22\%}$, which is still higher than Jain \etal \citet{jain2022distilling} by ${1.42\%}$. We refrain from reducing $n$ to $2$ as this would lead DIM to address only the single bias issue, failing to showcase our motivation and how DIM differs from Jain \etal \citet{jain2022distilling}.

\section{Experiments on Breeds}
\label{app:more_detail_breeds}
\subsection{Experimental Details}
We train ResNet-34 on the Breeds dataset from scratch. We use Adam as the optimizer. We set the learning rate as $0.01$ as the learning rate and the batch size as $32$.  We set the number of training epochs as $100$ and used early stop with validation loss as the criterion to eliminate the overfitting problem. 

\subsection{Subgroup Interpretation}

\begin{figure}
    \centering
    \includegraphics[width=\linewidth]{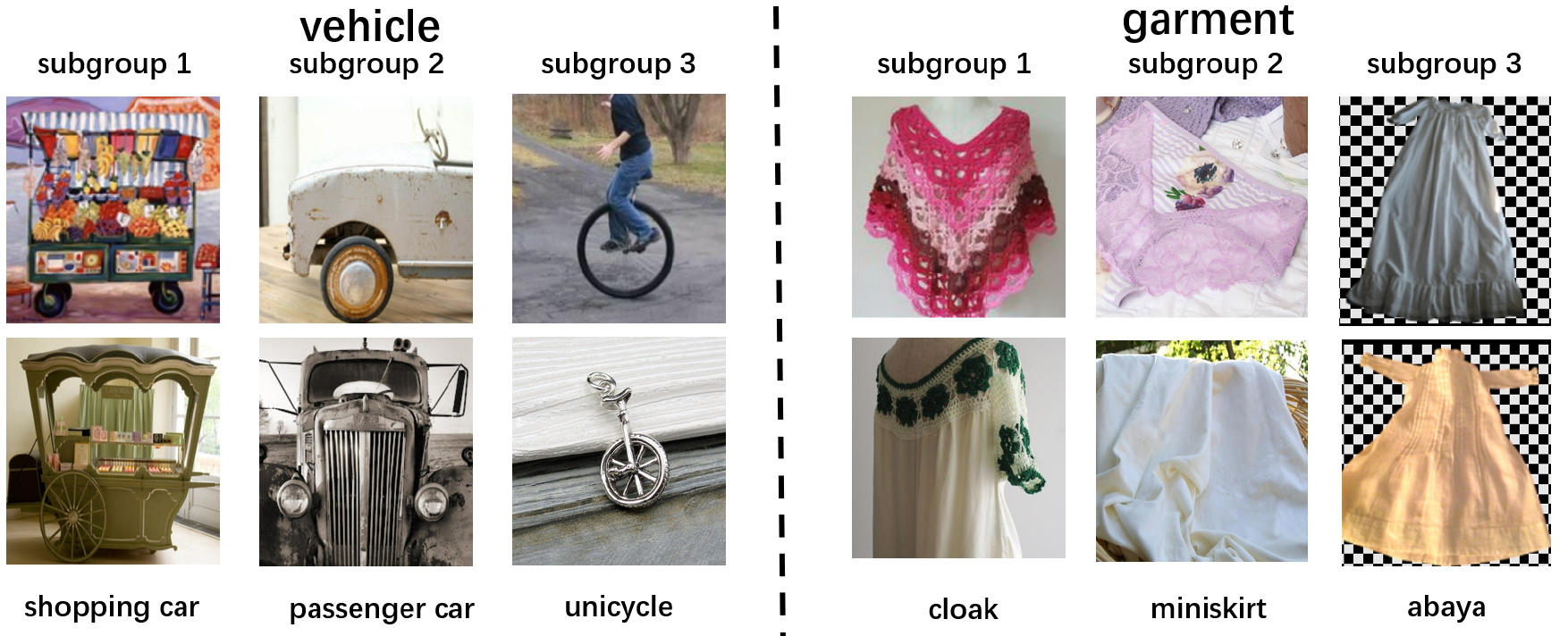}
    \caption{The CLIP Retrieval results of discovered subgroup embeddings in the class of ``vehicle'' and ``garment'' respectively. Images in each column come from the same identified subgroup. }
    \label{fig:breeds_1}
\end{figure}

\begin{figure}
    \centering
    \includegraphics[width=\linewidth]{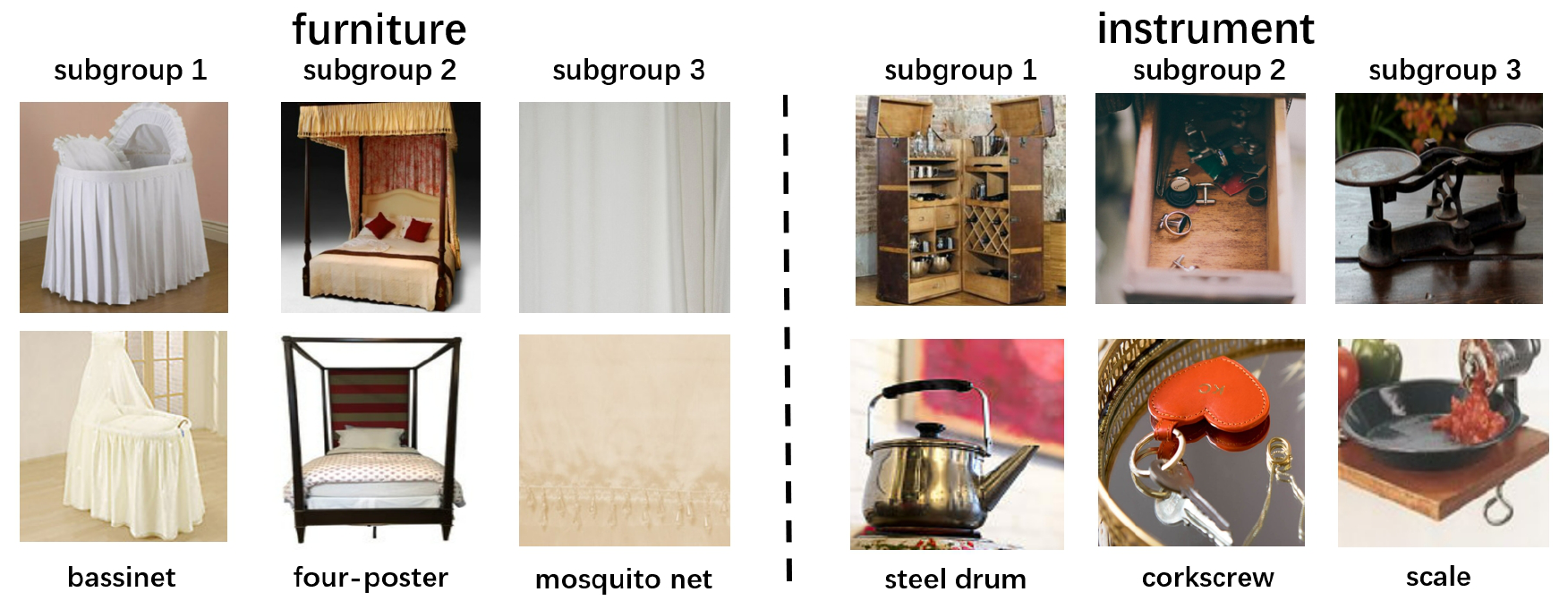}
    \caption{The CLIP Retrieval results of discovered subgroup embeddings in the class of ``furniture'' and ``instrument'' respectively. Images in each column come from the same identified subgroup. }
    \label{fig:breeds_2}
\end{figure}

We present several examples of discovering multiple subgroups in ResNet-34 on the Breeds dataset. As shown in \cref{fig:breeds_1}, our method can accurately capture multiple unknown subgroups, namely the ``shopping car'',  ``passenger car'', and ``unicycle'' in the ``vehicle'' class, ``cloak'', ``miniskirt'', and ``abaya'' in the ``garment'' class. Another confidence is provided in \cref{fig:breeds_2}. We can see that there are three\footnote{Here, we only present the results of the discovered three subgroups, while our method discovers $10$ subgroups in the experiments.} subgroups accurately discovered by our method, namely the ``bassinet'', ``four-poster'', and ``mosquito net'' in the ``furniture'' class, ``steel drum'', ``corkscrew'', and ``scale'' in the ``instrument'' class.

\section{Experiments on Hard ImageNet}
\label{app:more_detail_hard}
\subsection{Experimental Details}
We train ResNet-50 on the Hard ImageNet dataset. The ResNet-50 is pre-trained on the full ImageNet dataset and fine-tuned on the Hard ImageNet dataset. We use Adam as the optimizer. We set the learning rate as $0.01$ as the learning rate and the batch size as $32$.  We set the number of training epochs as $100$ and used early stop with validation loss as the criterion to eliminate the overfitting problem.

\subsection{Subgroup Interpretation}
\begin{figure}
    \centering
    \includegraphics[width=\linewidth]{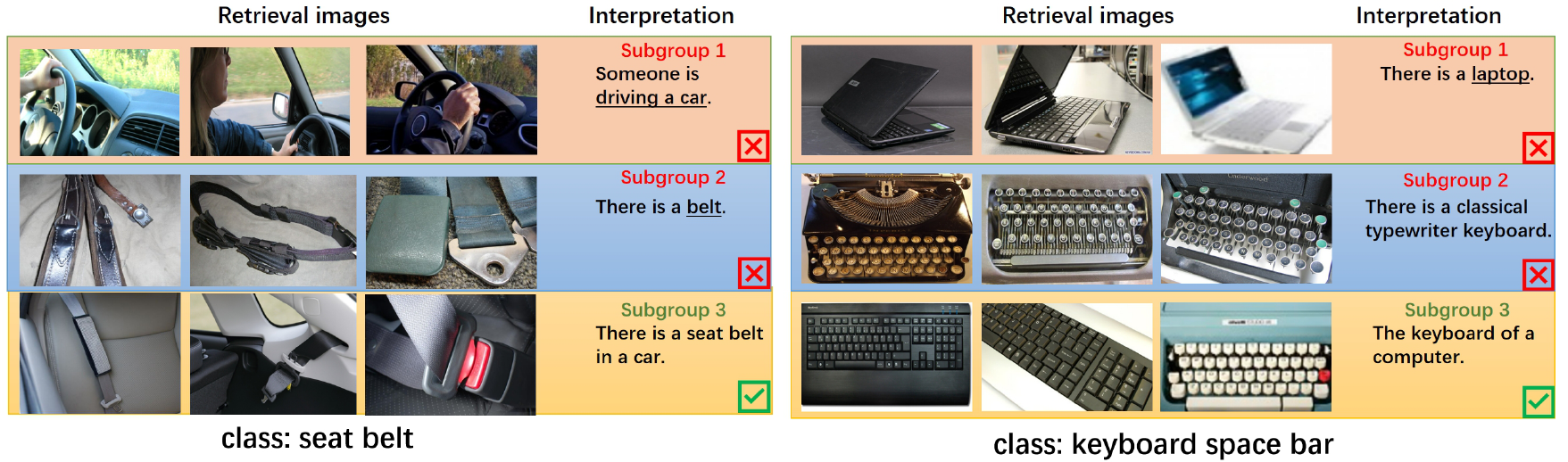}
    \caption{ Example of subgroup interpretation on Hard ImageNet. The first two rows are the retrieval images of identified biased subgroups
and corresponding summary descriptions by ChatGPT based on metadata. The last row is from the high-performance subgroup.}
    \label{fig:hard_supp_1}
\end{figure}

\begin{figure}
    \centering
    \includegraphics[width=\linewidth]{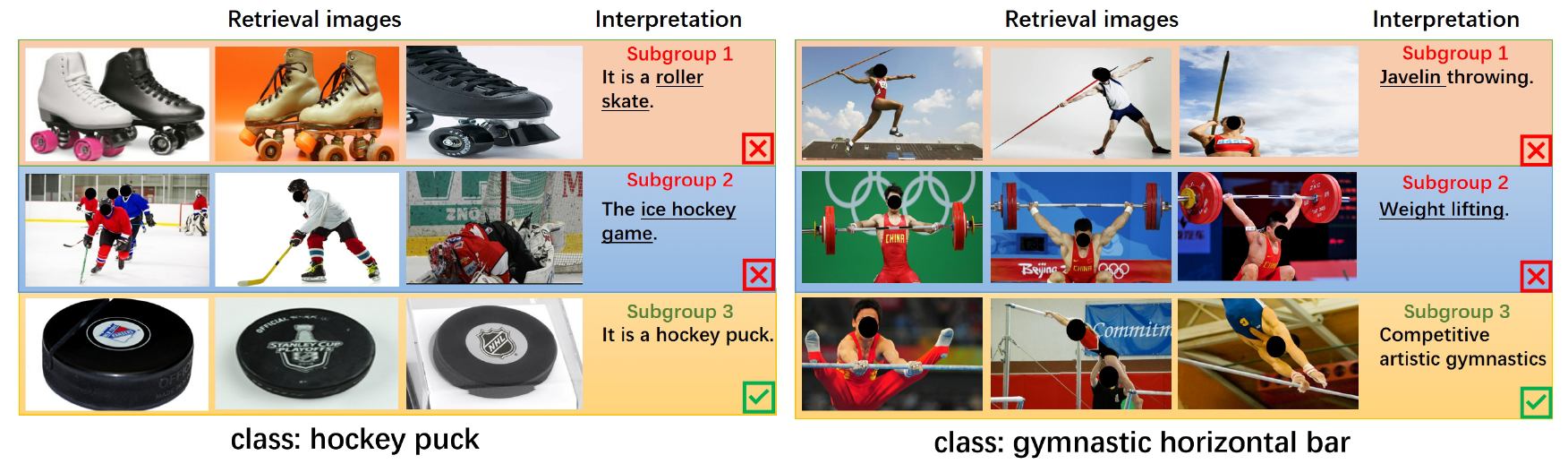}
    \caption{ Example of subgroup interpretation on Hard ImageNet. The first two rows are the retrieval images of identified biased subgroups
and corresponding summary descriptions by ChatGPT based on metadata. The last row is from the high-performance subgroup.}
    \label{fig:hard_supp_2}
\end{figure}

In \cref{fig: hard_imagenet_exp}, we uncover the implicit subgroups of the ``balance beam''  and ``dog sled'' classes in the Hard ImageNet dataset. Here, we additionally present four examples of the implicit multiple subgroups, namely the  ``seat belt'', ``keyboard space bar'', ``hockey puck'', and ``gymnastic horizontal bar'', in \cref{fig:hard_supp_1} and \cref{fig:hard_supp_2}, respectively. We provide the analysis of the spurious correlations involved in the dataset as follows. 

\noindent \textbf{``balance beam''}. The presented subgroups are ``a group of little kids playing'', ``the women's uneven bars event'', and ``'balance beam''. The model learns two biases, namely, the population and the scene. For the population, it means that the balance beam often comes up with kids, indicating a spurious correlation. For the scene, the balance beam is used for competition, causing unintended bias.

\noindent \textbf{``dog sled''}. The presented subgroups involve ``some people are skiing on the snow'', ``there are a lot of dogs'', and ``there are dog sleds''. Correspondingly, the results indicate the model unindently learns two biases. The first one comes from the background, where the dog sled usually appears in the snow. The second bias comes from the object, where the model is biased to the spurious correlation of dogs.

\noindent \textbf{``seat belt''}. The results of ``seat belt'' indicate that the model is largely biased by the background and unrelated objects, namely the car and belt, respectively.

\noindent \textbf{``keyboard space bar''}. The biases of the model on the ``keyboard space bar'' mainly derive from the objects, namely the laptop and the typewriter, which are the source of the keyboard.

\noindent \textbf{``hockey puck''}.  The biases of the model on the ``hockey puck'' mainly derive from the objects, namely the roller skate and ice hockey game, which usually come up with the hockey puck.

\noindent \textbf{``gymnastic horizontal bar''}. It is interesting to see the results of the subgroup discovery in the ``gymnastic horizontal bar''. It can be seen that the model detects the horizontal bar by the shape, leading to the wrong classification of the javelin and barbell.

\end{document}